%% file: main.tex
\documentclass[10pt,twocolumn,letterpaper]{article}

\usepackage{cvpr}              % To produce the CAMERA-READY version
\input{preamble}

\definecolor{cvprblue}{rgb}{0.21,0.49,0.74}
\usepackage[pagebackref,breaklinks,colorlinks,citecolor=cvprblue]{hyperref}
\usepackage{hyperref}       % hyperlinks
\usepackage{url}            % simple URL typesetting
\usepackage{booktabs}       % professional-quality tables
\usepackage{amsfonts}       % blackboard math symbols
\usepackage{nicefrac}       % compact symbols for 1/2, etc.
\usepackage{microtype}      % microtypography
\usepackage{lipsum}
\usepackage{fancyhdr}       % header
\usepackage{graphicx}       % graphics
\usepackage{multirow}
\usepackage{booktabs}
\usepackage{amsmath}
\usepackage{bm}
\usepackage{amssymb}
\usepackage{comment}
\usepackage{placeins}
\usepackage{algorithm}
\usepackage{algorithmic}
\usepackage[accsupp]{axessibility}

\def\paperID{8} % *** Enter the Paper ID here
\def\confName{CVPR}
\def\confYear{2024}

\title{StereoDiffusion: Training-Free Stereo Image Generation \\ Using Latent Diffusion Models}

\author{Lezhong Wang \quad Jeppe Revall Frisvad \quad Mark Bo Jensen \quad Siavash Arjomand Bigdeli\\ %
Technical University of Denmark, Kongens Lyngby\\
{\tt\small \{lewa, jerf, mboje, sarbi\}@dtu.dk}
}

\begin{document}
\twocolumn[{%
\renewcommand\twocolumn[1][]{#1}%

\maketitle
\begin{center}
    \centering
    \vspace{-2ex}
    \begin{minipage}[b]{0.02\textwidth}
    (a)\\[11ex]
    (b)\\[11ex]
    (c)\\[4ex]
    \end{minipage}
    \includegraphics[width=0.8\textwidth]{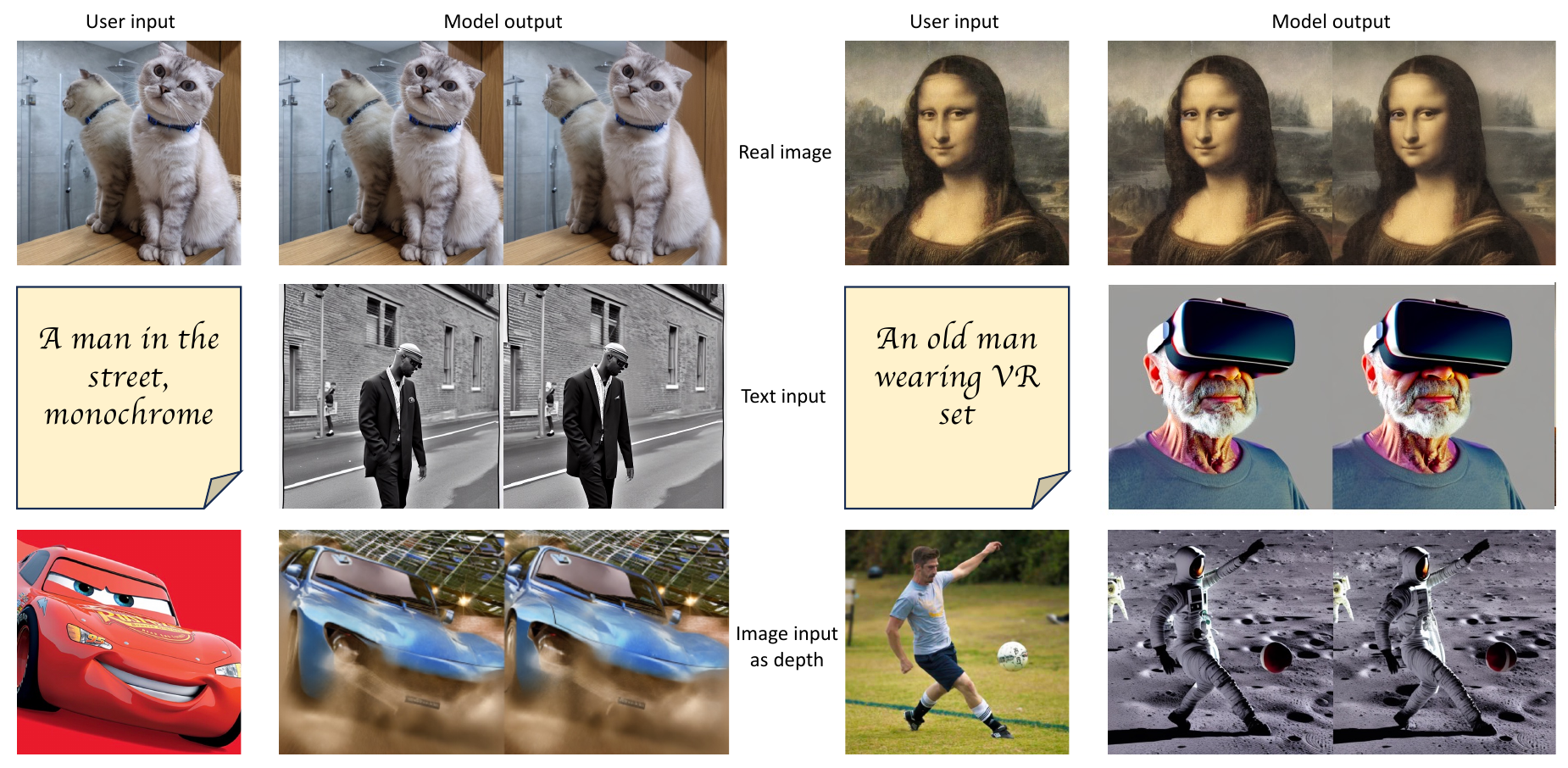}  \\[-2ex]
    \captionof{figure}{Our method takes one of three types of user input and generates a stereo image. Accepted user inputs: (a) a photo, (b) a text prompt, or (c) a user's image as a depth map and a prompt. We use a latent diffusion model pretrained on either images (a, b) or depth maps (c).}
    \label{fig:teaser}
\end{center}%
}]

% \begin{figure*}
%   \centering
%   \includegraphics[width=0.78\textwidth]{Figs/cover.pdf}  \\[-2.5ex]
%   \caption{Our approach takes one of three types of user input and generates a stereo image. The accepted user inputs are (a) a photo, (b) a text prompt, or (c) a user's image as a depth map and a prompt. We use a latent diffusion model pretrained on images for inputs a and b and on depth maps for input c.}
% \end{figure*}
%\maketitle
\input{sec/0_abstract}    
\input{sec/1_body}

% \clearpage
% \input{sec/2_formatting}
% \input{sec/3_finalcopy}
{
    \small
    \bibliographystyle{ieeenat_fullname}
    \bibliography{references}
}

\input{sec/X_suppl}

\end{document}

%% file: preamble.tex
\usepackage[dvipsnames]{xcolor}

%% file: sec/0_abstract.tex
\begin{abstract}
The demand for stereo images increases as manufacturers launch more extended reality (XR) devices. To meet this demand, we introduce StereoDiffusion, a method that, unlike traditional inpainting pipelines, is training-free and straightforward to use with seamless integration into the original Stable Diffusion model. Our method modifies the latent variable to provide an end-to-end, lightweight method for fast generation of stereo image pairs, without the need for fine-tuning model weights or any post-processing of images. Using the original input to generate a left image and estimate a disparity map for it, we generate the latent vector for the right image through Stereo Pixel Shift operations, complemented by Symmetric Pixel Shift Masking Denoise and Self-Attention Layer Modifications to align the right-side image with the left-side image. Moreover, our proposed method maintains a high standard of image quality throughout the stereo generation process, achieving state-of-the-art scores in various quantitative evaluations.
\end{abstract}

%% file: sec/1_body.tex
\section{Introduction}

Large-scale language-image (LLI) models have become prominent in recent years, acclaimed for their advanced generative semantic and compositional abilities~\cite{saharia2022photorealistic,ramesh2022hierarchical,yu2022scaling,nichol2021glide,ding2022cogview2}. Their distinctiveness lies in their training on extensive language-image datasets, enabling them to interpret and generate content from diverse linguistic and visual contexts. Utilizing innovative image generative techniques such as auto-regressive and diffusion models~\cite{hertz2022prompt} have significantly advanced the synergy between linguistic understanding and image generation. This has led to a new era of creative and semantically rich image synthesis, marking notable advancements in artificial intelligence and computer vision.

A significant recent development in the VR/AR field is Apple's introduction of Vision Pro, which has the potential to drive rapid advancements in this field.
Despite the growing production of 3D content by various manufacturers, and related research \cite{liu2023one,wu2023multiview,li20233dqd,wu2023aniportraitgan,rockwell2021pixelsynth} in recent years, the availability of stereo multimedia content, which offers a depth-enhanced visual experience, remains relatively scarce. As the VR/AR era looms, the limitations of existing image generation models that are confined to producing 2D images become increasingly apparent. However, there is currently no relevant research that attempts to use image generation models to directly generate stereo image pairs.
In response to this challenge, we introduce a novel methodology. Through modification of the Stable Diffusion model's latent variable, we have devised an efficient end-to-end approach, eliminating the need for additional models like inpainting \cite{shih20203d,hu2021worldsheet} for post-processing to generate stereo images. Figure~\ref{fig:teaser} presents some examples. Code is available at \href{https://github.com/lez-s/StereoDiffusion}{https://github.com/lez-s/StereoDiffusion}.

\begin{figure*}[tbp]
  \centering
  % \mbox{} \hfill
  \includegraphics[width=0.9\linewidth]{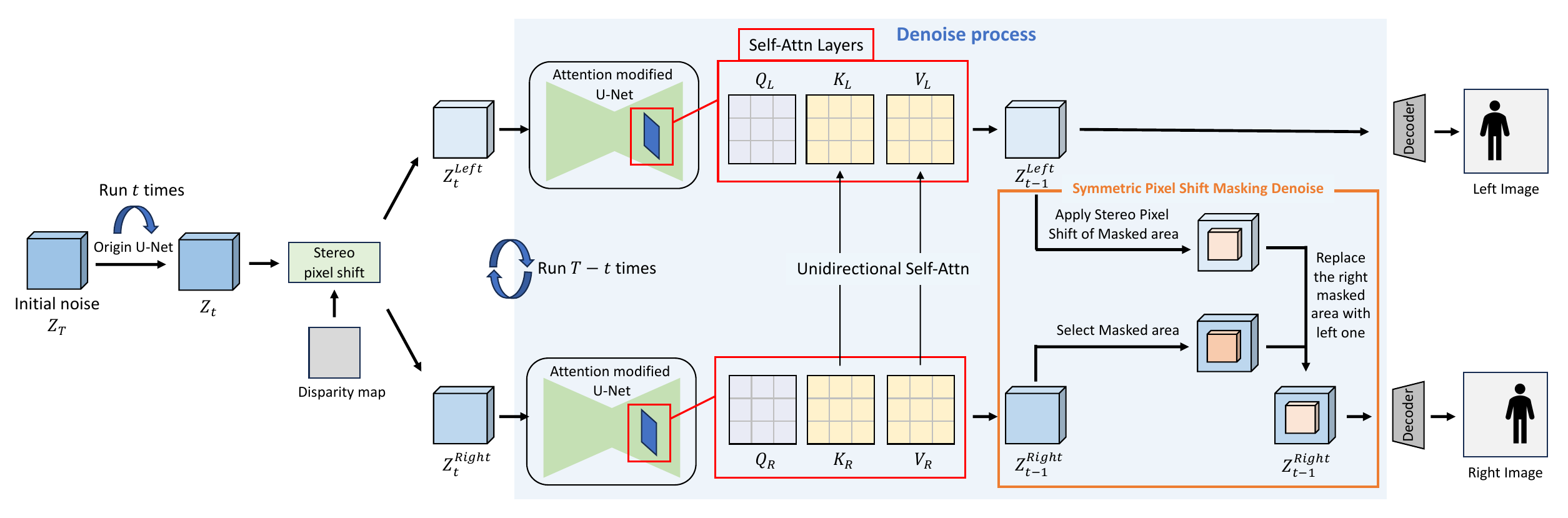} \\[-2.5ex]
  \caption{The pipeline of our Stereo Diffusion. The process starts with random noise and denoising of it to generate a stereo image pair. The operation of Stereo Pixel Shift is represented by Eq.~\ref{eq:sps}. The Disparity Map for generating stereo image pairs can be obtained from depth models such as DPT~\cite{Ranftl2021} or MiDas~\cite{Ranftl2022}. The pipeline only shows the Unidirectional Self-Attention operation, designed to align the right-side image with the left-side image, a method that satisfies general needs. Bidirectional Self-Attention, being a mutual operation, would be represented by bidirectional arrows in the image. The orange box in the image depicts the concept of Symmetric Pixel Shift Masking Denoise, with details explained in Sec.~\ref{sec:spsmd}. The cross attention part of the sampling process is omitted for brevity.}
  \label{fig:pipeline}
\end{figure*}

We address the constraints of traditional image generation models that employ an inpainting pipeline. Our approach is to generate stereo image pairs by adjusting the latent variable of the Stable Diffusion model, see Figure~\ref{fig:pipeline}. We use Symmetric Pixel Shift Masking Denoise and Self-Attention layer modifications to align the generated right-side image with the left-side image. This method allows for a lightweight solution that can be seamlessly integrated into the original Stable Diffusion model without the need for fine-tuning. To the best of our knowledge, our approach represents the first instance of generating stereo images by modifying the latent variable of Stable Diffusion. Compared with other methods, our approach enables the training-free end-to-end rapid generation of high-quality stereo images using only the original Stable Diffusion model.

%-------------------------------------------------------------------------

\section{Related work}

\iffalse
\paragraph*{Latent space of Generative Adversarial Networks.} The evolution of image generation models, notably starting from Generative Adversarial Networks (GANs)~\cite{reed2016generative,radford2016unsupervised,isola2018imagetoimage,gulrajani2017improved}, has ushered in an era of remarkable progress in generating highly realistic images.  One of the pivotal attributes of GANs is their latent space, demonstrating a robust correlation with the generated images~\cite{ramesh2022hierarchical,rombach2022high,pan2023drag}. 
This characteristic allows modification of the generated images by manipulating the latent space.
The intrinsic relationship between the latent space and the resultant images has catalyzed extensive research endeavors, particularly focusing on the latent space of GANs ~\cite{abdal2021styleflow,leimkuhler2021freestylegan,patashnik2021styleclip,shen2020interpreting,edo2020editing}. These studies have explored diverse applications, ranging from sophisticated image editing techniques to intricate inversion methodologies~\cite{bermano2022state,xia2022gan}. A plethora of research has delved into the intricacies of GAN latent spaces, unveiling their potential in shaping the outcomes of image generation processes. Notably, researchers have probed the manipulation of GAN latent spaces~\cite{pan2023drag}, revealing that precise control over the image generation process is attainable through interventions in the latent space.
\fi

\paragraph*{Latent space of a Latent Diffusion Model.} 
Diffusion models, notably the Denoising Diffusion Implicit Models (DDIM)~\cite{song2020denoising}, have made significant strides in image generation. The DDIM sampling algorithm revealed that using the same initial noise results in consistent high-level features across different generative paths, highlighting initial noise as a potent latent image encoding~\cite{song2020denoising}. This discovery aids in modifying images by adjusting the Stable Diffusion latent variable. A key challenge in stereo image generation is maintaining content consistency between paired images. Researchers are focusing on image editing techniques using Stable Diffusion~\cite{cao2023masactrl,tumanyan2023plug,parmar2023zero,yang2022diffusion}, such as the "prompt-to-prompt" method~\cite{hertz2022prompt}, which involves altering the model's cross-attention during sampling for text-prompt-based image editing. 
ControlNet~\cite{zhang2023adding} is also a notable work in the field, but instead of utilizing the latent space, the authors trained ControlNet on a large dataset to better control the generation of desired images by Stable Diffusion. Additionally, ControlNet primarily focuses on pose control and lacks the capability for pixel-level modifications of images.
Although effective, these methods are less suited for tasks needing precise pixel-level manipulation, like stereo image generation, due to their reliance on text prompts for image modification.

\paragraph*{Video generation by Latent Diffusion Model.} Ensuring the consistency of images within the same batch in Stable Diffusion has long been a challenge in video generation~\cite{wang2023videocomposer,blattmann2023align}. VideoComposer addresses this by incorporating an STC-encoder into the Latent Diffusion Model's U-Net architecture, ensuring consistency in the generated image content~\cite{wang2023videocomposer}. Similarly, VideoLDM achieves impressive video generation results by introducing 3D convolution layers and temporal attention layers into the spatial and temporal layers of U-Net~\cite{blattmann2023align}. However, these methods require fine-tuning of the original models and substantial amounts of data. Generally, this is not an issue for video generation, but for achieving stereo image generation, the available stereo image data is quite limited, mostly comprising road traffic images initially intended for autonomous driving depth prediction services. In the video generation field, attempts such as Tune-A-Video~\cite{wu2022tune} have explored zero-shot video generation~\cite{wu2022tune,wang2023zero}. This work utilizes a technique called ST-Attn to maintain the continuity of videos. We employ a comparable approach to ensure consistency between the left and right images.

\paragraph*{3D photography and inpainting.} Traditional image-based reconstruction and rendering methods require complex capture setups, involving numerous images with significant baselines~\cite{hedman2017casual,whelan2018reconstructing,hedman2018deep,kopf2013image}. Currently, limited research endeavors directly focus on generating stereo images. Many studies have concentrated on generating 3D photos, a technique allowing subtle changes in the camera angle for observing photos from different perspectives~\cite{hu2021worldsheet,shih20203d,zhou2018stereo,srinivasan2019pushing,mildenhall2019local}. Among 3D image generation techniques, 3D Photography Inpainting is a notable approach \cite{shih20203d,hedman2017casual}. This method employs inpainting to generate 3D images. After passing the input image through a depth estimation model, they map the image onto a mesh and apply changes in perspective based on the depth map of the original image. Inpainting is then utilized to fill the gaps left by transformed pixels in the original image. This approach significantly differs from our modification of the Stable Diffusion latent space. Although this method could be adopted as post-processing of an image generated through Stable Diffusion, it requires additional steps and consumes more time.

\paragraph*{3D scene generation by pretrained Stable Diffusion.}
Recent studies have used model distillation with pretrained Stable Diffusion models for text-based 3D reconstruction. DreamFusion~\cite{poole2022dreamfusion} employes `Score Distillation Sampling' (SDS) to initialize and render a NeRF model, improving with Imagen-surrounding score distillation loss. Variational Score Distillation (VSD)~\cite{wang2023prolificdreamer} further enhances 3D scene quality. These methods can create stereo images via rendering but are time-intensive. Without full 3D scenes, our method provides a faster solution for generating stereo image pairs.

\section{Method}
Diverging from conventional inpainting methods, our approach is distinctively simple and training-free. With seamless integration into the original Stable Diffusion framework, we provide end-to-end generation of stereo image pairs, eliminating the need for post-processing. Our method leverages a disparity map in the early denoising stage to apply a Stereo Pixel Shift (Sec.~\ref{sec:sps}) to the latent vector of the left image. This process generates the latent vector for the right image through disparity. To address the inconsistency issues between the left and right images during the denoising process, we employ a Symmetric Pixel Shift Masking Denoise (Sec.~\ref{sec:spsmd}) technique and a Self-Attention module (Sec.~\ref{sec:salm}) to align the right image with the left one. 
Since our method exclusively manipulates the latent variable, it can be applied across various image generation tasks in different Stable Diffusion models. This versatility stems from the technique's focus on latent space operations, making it adaptable to a wide range of scenarios within the Stable Diffusion framework (Sec.~\ref{sec:as}). Our method only requires a disparity map, which can be obtained by various depth estimation models like DPT~\cite{Ranftl2021}, MiDas~\cite{Ranftl2022} etc.\ and does not require camera calibration.

\subsection{Stereo Pixels Shift} \label{sec:sps}
For the task of generating stereo images, fine-tuning models on large stereo datasets like KITTI~\cite{menze2015object} seems intuitive. However, after fine-tuning the model using various methods such as ControlNet~\cite{chen2023control} and LoRA~\cite{hu2021lora}, the results of the generated images remains unsatisfactory. A major flaw of this approach is that even if we could generate a high-quality stereo image pairs, the types of images generated will be limited to driving scenes similar to KITTI, losing the most important feature of Stable Diffusion: its diversity. Inspired by the Denoising Diffusion Implicit Models (DDIM) sampling technique for Stable Diffusion~\cite{song2020denoising}, we present a new method, Stereo Pixels Shift, without the aforementioned drawbacks. 
Utilizing DDIM for sampling from generalized generative processes, a latent vector sample $\bm{x}_{t-1}$ is generated from a sample $\bm{x}_t$ via a noise predictor $\epsilon_\theta$:
\begin{equation}
\begin{aligned}
\bm{x}_{t-1} =\: &\sqrt{\alpha_{t-1}}\, \underbrace{\left(\frac{\bm{x}_t - \sqrt{1-\alpha_t} \, \epsilon_\theta^{(t)}\!\!\left(\bm{x}_t\right)}{\sqrt{\alpha_t}}\right)}_{\text{predicted } \bm{x}_0 }  \\
&+ \underbrace{\sqrt{1-\alpha_{t-1} - \sigma_t^2} \, \epsilon_\theta^{(t)}\!\!\left(\bm{x}_t\right)}_{\text{direction pointing to } \bm{x}_t } + \underbrace{\sigma_t \epsilon_t}_{\text{random noise}} \,,
\end{aligned}
\label{math:ddim}
\end{equation}
where $\epsilon_t$ is noise following a standard Gaussian distribution $\mathcal{N}(\mathbf{0}, \mathbf{I})$, independent of $\bm{x}_t$, and $\alpha_t$ controls the noise scale at step $t$ with $\alpha_0 := 1$. If we set $\sigma_t = 0$ for all $t$ and the same model $\epsilon_{\theta}$ is used, the generative results are consistent and identical, making the forward process deterministic, given $\bm{x}_{t-1}$ and $\bm{x}_0$. Thus, the result of $\bm{x}_{t-1}$ depends solely on \( \bm{x}_{t} \). During the denoise process at a certain step \( t' \), if we modify \( \bm{x}_{t'} \) to \( \bm{x}'_{t'} \), subsequently, \( \bm{x}'_{t'-1} \) is denoised based on \( \bm{x}'_{t'} \), eventually generating \( \bm{x}'_{0} \) which is different from the original \( \bm{x}_{0} \).
This pivotal insight enables the practical application of Stable Diffusion for stereo image generation. To align with this approach, we scale down the disparity map to match the dimensions of the latent space. Subsequently, we manipulate the latent vector on a pixel-by-pixel basis, guided by the disparity map. Given the relatively small size of the latent vector, this process does not entail a substantial computational overhead. 

Assuming that the two images have parallel optical axes, we derive a disparity map from a depth map using
\begin{equation}
    D(x,y) = \frac{fB}{Z(x, y)} \,,
\end{equation}
where $(x,y)$ is a point in image space, $Z$ is the depth map, $f$ represents the focal length, and $B$ is the baseline distance (i.e., the distance between the two cameras). Typically, we normalize the range of the disparity map $D(x,y)$ to be in $[0,1]$. When the disparity map is generated by a model rather than being measured by actual devices, the conversion process is unnecessary, since many depth estimation models are capable of directly generating disparity maps.

The Stereo Pixel Shift operation $\mathcal{S}$ is expressed by
\begin{equation}
    \mathbf{x}_{\text{right}}(x, y) = \mathbf{x}_{\text{left}}(x - s\, D(x, y), y) \,,
    \label{eq:sps}
\end{equation}
where $\mathbf{x}$ (left or right) denotes the latent variable $\bm{x}_t$, $\mathbf{x}_{\text{left}}(x - s \, D(x, y), y)$ represents the position in the latent space that is shifted left by $D(x, y)$ pixels relative to the position $(x, y)$ in the latent space, and $s$ is a scaling factor that controls the range of disparity, i.e., the pixel shift distance of the point closest to the observer in the right image relative to the left. Within reasonable limits, a larger value of $ s $ enhances the stereo effect of the generated images, usually restricted to within 10\% of the image width. Excessively large $s$ values can cause discomfort or blurriness rather than a sense of depth.
However, using this method on images directly can lead to problems like flying pixels, as it causes individual pixels to warp into the empty spaces between two depth surfaces~\cite{watson2020learning}. But since we operate on pixels in the latent space, individual pixel issues are typically resolved in the subsequent denoising and decoding processes. Thus, our method is straightforward, requiring no additional processing such as sharpening of the moved pixels. 

\begin{figure}
    \centering
    \includegraphics[width=\linewidth]{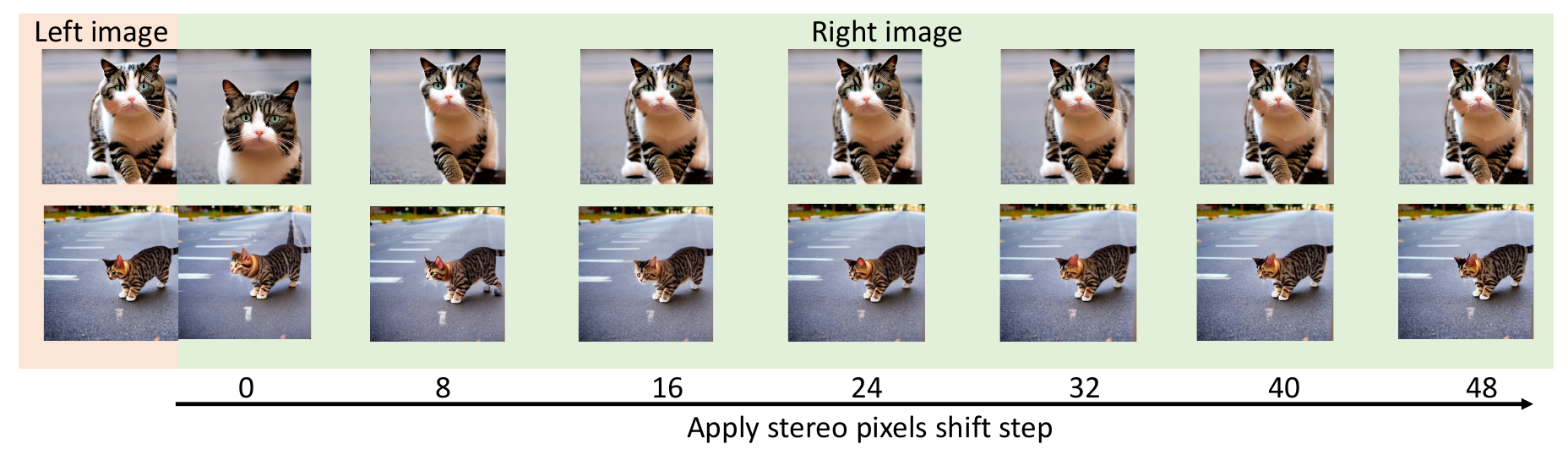} \\[-2ex]
    \caption{Comparing the outcomes of applying stereo shifts at different steps of denoising, reveals varying optimal configurations for different images. Implementing shifts too early could result in significant content alterations, while shifts applied too late might lead to noticeable artifacts in the images.}
    \label{fig:stereoshift}
\end{figure}

The reason why we can apply Stereo Pixel Shift to the latent variable is that, after a certain step, there is a spatial position correspondence between the latent variable and the generated image.
According to diffusion process theory~\cite{song2021scorebased,preechakul2021diffusion,song2020denoising,yang2022diffusion,nichol2021improved,bao2022analyticdpm}, sampling can be represented as
\begin{equation}
\bm{x}_t=\sqrt{\bar{\rule{0pt}{1.18ex}\alpha}_t} \bm{x}_0+\sqrt{ \left(1-\bar{\rule{0pt}{1.18ex}\alpha}_t\right)} \epsilon \,,
\end{equation}
where $\epsilon \sim \mathcal{N}(\mathbf{0}, \mathbf{I})$ and $\bar{\alpha}_t=\prod_{i=1}^t \alpha_i$. Applying the Fourier transform on both sides, we have
\begin{equation}
\mathcal{F}\!\left(\bm{x}_t\right)=\sqrt{\bar{\rule{0pt}{1.18ex}\alpha}_t}\, \mathcal{F}\!\left(\bm{x}_0\right)+\sqrt{\left(1-\bar{\rule{0pt}{1.18ex}\alpha}_t\right)} \,\mathcal{F}(\epsilon) \,.
\end{equation}
If the step $t$ is small, then $\bar{\rule{0pt}{1.18ex}\alpha}_t \approx 1$, which indicates that early-stage sampling involves low-frequency signals primarily defining the contours of the generated image. When the step $t$ is large, $\bar{\rule{0pt}{1.18ex}\alpha}_t \approx 0$, high-frequency signals in the later-stage sampling refine the image details. This results in a significant disparity between the generated image and the original image if pixel offsets are applied too early during the sampling steps. Applying pixels shifts too late maintains high consistency in image content, but results in noticeable artifacts in the generated images. We found through experiments that setting $t$ to 20\% of the total denoise step usually works well.
Additionally, the appropriate sampling steps for pixel offsets vary depending on the size of the objects in the images, see Figure~\ref{fig:stereoshift}. 

\subsection{Symmetric Pixel Shift Masking Denoise} \label{sec:spsmd}

After applying the Stereo Pixel Shift, the right latent vector becomes inconsistent with the left one, potentially leading to discrepancies in the moved subject content following the denoising process. 

According to Eq.~\ref{math:ddim}, a pixel shift applied to $ x_{t'} $ to obtain $ x'_{t'} $ results in a slight difference between $ x'_{t'-1} $ and $ x_{t'-1} $. This difference accumulates during the subsequent denoising process, leading to variations in the final generated image. As a result, when the denoising algorithm is applied, it may interpret the shifted areas differently, potentially causing variations in how the subject matter appears after processing. This challenge is crucial in stereo image generation, as maintaining symmetry and coherence between the two sides is essential for creating a convincing and realistic stereo effect.
 
To circumvent the issue, inspired by the concept of inpainting, we propose a Symmetric Pixel Shift Masking Denoise method. 
We create a mask for the area where the stereo pixel shift is applied. At regular intervals, defined by specific steps $t'$, the values from the masked region of the left latent space are copied to the corresponding area of the mask in the right latent space. Consequently, the denoising process for the right image can be reformulated from Eq.~\ref{math:ddim} as
\begin{equation}
\begin{aligned}
\bm{x}'_{t'-1} =\: &\sqrt{\alpha_{t'-1}}\, {\left(\frac{\bm{X}'_{t'} - \sqrt{1-\alpha_{t'}}\, \epsilon_\theta^{(t')}\!\left(\bm{x}'_{t'}\right)}{\sqrt{\alpha_{t'}}}\right)}  \\
&+ \sqrt{1-\alpha_{t'-1}} \, \epsilon_\theta^{(t')}\!\left(\bm{x}'_{t'}\right) ,
\end{aligned}
\end{equation}
where $ \bm{x}'_{t'} $ represents the right latent vector after undergoing a pixel shift, and the $i$th element of $\bm{X}'_{t'}$ is expressed by
\begin{equation}
\bm{X}'_{t',i} = 
  \begin{cases} 
   \mathcal{S}(\bm{x}_{t'-1, i},D) & \text{if } \mathbf{M}_i = \text{True}, \\
   \bm{x}'_{t', i}       & \text{otherwise,}
  \end{cases}
  \label{eq:Xt}
\end{equation} 
where $\mathcal{S}$ represents the operation of Stereo Pixel Shift in Eq.~\ref{eq:sps}, $D$ is the corresponding disparity map of the image, and $ \mathbf{M} $ is a Boolean matrix of the same shape as $\bm{x}$ that signifies the mask, with values set to True for the pixels that have been shifted. The variable $\bm{x}_{t'-1}$ denotes the latent vector of the left image at time step $t'-1$.

We note that if the area shifted is left blank (i.e., filled with zeros), the denoised region might become blurry. We address this blurriness by filling the shifted blank area with random noise using
\begin{equation}
\bm{x}^{(\text{deblur})}_{t',i} = 
  \begin{cases} 
   \bm{x}_{t',i} & \text{if } \mathbf{M}_i = \text{False}, \\
   \epsilon_{t',i}       & \text{otherwise,}
  \end{cases}
\end{equation} 
where $\epsilon_{t'}$ denotes random noise, $\mathbf{M}$ is the same mask as the one in Eq.~\ref{eq:Xt}. However, the effectiveness varies with different images. Sometimes, it may even lead to a decrease in the quality of the generated images. A detailed effects analysis of the Deblur technique is presented in the ablation studies described in Sec.~\ref{sect:ablation}.

\subsection{Self-Attention layers modification} \label{sec:salm}
As numerous studies have attempted to modify the attention mechanisms within Stable Diffusion to achieve the goal of modifying the original images~\cite{cao2023masactrl,hertz2022prompt,wang2023zero,parmar2023zero,wu2022tune}, we tackle this challenge by utilizing both Unidirectional and Bidirectional Self-Attention mechanisms. This method eliminates the need for fine-tuning the model to adjust its weights. Refer to supplementary materials for detailed explanation.
\begin{algorithm} []
    \caption{Bi/Uni-directional Attention Modification}

        \begin{algorithmic}[1]
        \REQUIRE A text condition $\mathcal{C}$, a left latent variable $z_{t-1}$ and a right latent variable $z'_{t-1}$.\\
        \ENSURE An edited right latent variable $z'^*_{t-1}$ and an edited latent latent variable $z^*_{t-1}$ if bidirection.
            \STATE $(z_{t-1},z'_{t-1}),(M_t,M'_t) \leftarrow \epsilon_\theta((z_t,z'_t),t,\mathcal{C}) $;
            \STATE $\widehat{M}_t,\widehat{M}'_t \leftarrow \operatorname{Edit}\left(M_t, M'_t, t\right)$ ;
            \IF{Unidirection}
                \STATE $ (z_{t-1},z'^*_{t-1}) \leftarrow \epsilon_\theta((z_t,{z'_t}),t,\mathcal{C})\{ M' \leftarrow \widehat{M'_t}\}$
                \RETURN $(z_{t-1},z'^*_{t-1})$
            \ELSIF{Bidirection}
                \STATE $ (z^*_{t-1},z'^*_{t-1}) \leftarrow \epsilon_\theta((z_t,{z'_t}),t,\mathcal{C})\{ M \leftarrow \widehat{M_t},M' \leftarrow \widehat{M'_t} \}$
                \RETURN $(z^*_{t-1}, z'^*_{t-1})$
            \ENDIF

        \end{algorithmic}
        \label{alg}
\end{algorithm}

Our modified approach is listed in Algorithm~\ref{alg}.
The term $\epsilon_\theta((z_t,z'_t),t,\mathcal{C})$ represents the computation of a single step $t$ of the diffusion process, which yields the noisy image $z_{t-1}$ and the attention map $M_t$. Here, $(z_t,z'_t)$ denote the left and right latent variables, respectively. In our implementation, these latent variables are stacked together along the batch size dimension. However, they are presented separately here for ease of explanation.
The expression $\epsilon_\theta((z_t,{z'_t}),t,\mathcal{C})\{ M' \leftarrow \widehat{M'_t}\}$ denotes the diffusion step where the attention map $M$ is superseded by an additional map $\widehat{M}$. 
We define the function $\operatorname{Edit}\left(M_t, M'_t, t\right)$ as a general edit function, designed to process the $t$-th attention maps of the left and right latent variables.

\begin{figure*}[htbp]
    \centering
    \includegraphics[width=0.9\linewidth]{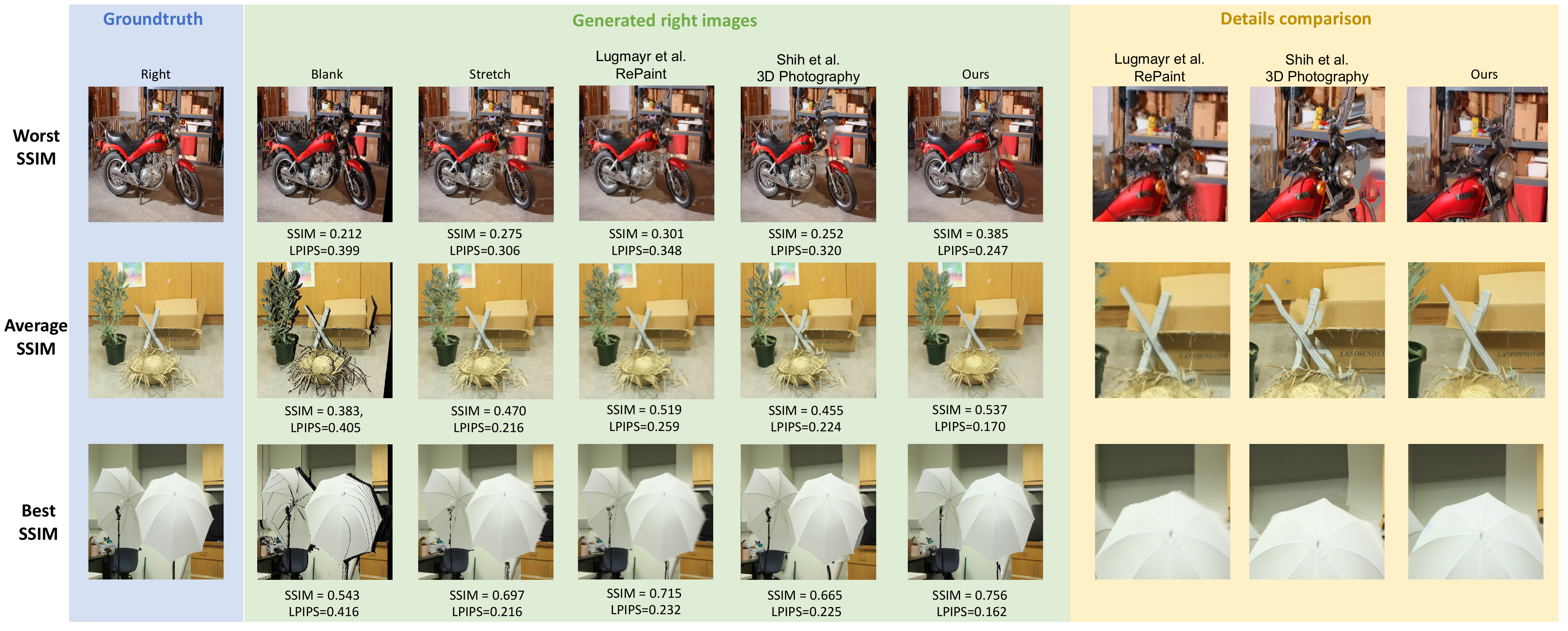} \\[-2ex]
    \caption{
    The same image generated using different methods. The rows present, respectively, the image with the lowest (worst) SSIM score, the image closest to the average SSIM score, and the image with the highest (best) SSIM score generated using our method. The other methods are represented solely by their results on these specific images, which do not necessarily reflect the best, average, or worst SSIM scores achievable by those methods. We do this to facilitate a direct comparison of the effects of each method on the same image. We also provide LPIPS scores for reference and close-ups of the images generated by the primary benchmark methods for inspection of details. %A `Details Comparison' is provided for a detailed comparison of images generated by the primary benchmark methods.
    }
    \label{fig:comp_ssim}
\end{figure*}

\begin{table*}
    \centering
    \caption{Quantitative evaluation results for Middlebury and KITTI: 
    the results of generating right-side images from left-side images and disparity maps using different methods. We assess the similarity between the generated and the original images using PSNR, SSIM, and LPIPS. `GT' indicates the use of ground truth disparity maps, while `pseudo' denotes the use of disparity maps generated by a depth estimation model. Scores presented in bold indicate the best performance. The numbers in the top right represent the best scores, while those in the bottom right indicate the worst scores. }
    \label{tab:qe_mb}
    \vspace{-1.5ex}
    {\small
    \begin{tabular}{c|ccc|ccc}
        \hline
        \multirow{2}{*}{Methods} & \multicolumn{3}{c|}{\rule{0pt}{2.5ex}Middlebury}& \multicolumn{3}{c}{KITTI} \\
        \cline{2-7}
        & \rule{0pt}{2.5ex}PSNR $\uparrow$ & SSIM $\uparrow$& LPIPS $\downarrow$&  PSNR $\uparrow$& SSIM $\uparrow$& LPIPS $\downarrow$\\
        \hline
        Leave blank\rule{0pt}{3ex} & $11.328^{+2.483}_{-3.489}$& $0.315^{+0.230}_{-0.154}$ &$0.450^{-0.089}_{+0.097}$&$12.980^{+4.919}_{-3.831}$&$0.374^{+0.286}_{-0.251}$&$0.313^{-0.109}_{+0.222}$ \\ [3pt]
        Stretch & $14.842^{+2.714}_{-2.753}$ &$0.432^{+0.265}_{-0.190}$ &$0.285^{-0.089}_{+0.112}$&$14.757^{+5.375}_{-4.694}$ &$0.429^{+0.287}_{-0.271}$&$0.212^{-0.100}_{+0.145}$\\[3pt]
        3D Photography~\cite{shih20203d} & $14.190^{+2.464}_{-2.798}$&$0.427^{+0.238}_{-0.175}$&$0.275^{-0.065}_{+0.073}$ &$14.540^{+7.256}_{-4.023}$ &$0.398^{+0.270}_{-0.323}$ &$0.210^{-0.073}_{+0.099}$\\[3pt]
        RePaint~\cite{lugmayr2022repaint} & $15.102^{+2.909}_{-2.802}$&$0.462^{+0.253}_{-0.184}$&$0.311^{-0.079}_{+0.090} $&$15.056^{+5.366}_{-4.897}$&$0.462^{+0.268}_{-0.285}$&$0.251^{-0.095}_{+0.128}$ \\[3pt]
        Ours (with GT disparity) & $15.456^{+2.669}_{-3.313}$&$0.468^{+0.252}_{-0.205}$&$0.231^{-0.088}_{+0.096}$ &$\textbf{15.679}^{+5.888}_{-5.487}$ &$\textbf{0.481}^{+0.245}_{-0.310}$ &$\textbf{0.205}^{-0.099}_{+0.135}$\\[3pt]
        Ours (with pseudo disparity)&$\textbf{16.980}^{+4.737}_{-3.818}$&$\textbf{0.551}^{+0.208}_{-0.166}$ &$\textbf{0.173}^{-0.069}_{+0.074}$ &$15.589^{+8.061}_{-5.016}$&$0.479^{+0.241}_{-0.300}$&$0.209^{-0.116}_{+0.114}$\\[1ex]
        \hline
        \multicolumn{7}{c}{}\\[-5.5ex]
    \end{tabular}}
    %\\[-1ex]
\end{table*}

\subsection{Application scenarios} \label{sec:as}
As shown in Figure~\ref{fig:teaser}, our method is compatible with various types of Stable Diffusion models, enabling it to: (a) produce the corresponding right-side image from an existing left-side image; (b) generate stereo images from text prompts; (c) produce the corresponding right-side image from an existing left-side image, where the pair shares the same composition but differs in content. For text-to-image and depth-to-image tasks, the initial noise is randomly generated. Thus, it is sufficient to apply a pixel shift to the denoised noise after a specific denoising step, as illustrated in Figure~\ref{fig:pipeline}. However, for generating stereo image pairs of an existing image, it is necessary to use null-text inversion~\cite{mokady2023null} to obtain the latent space of the original image. A straightforward inversion technique was proposed for DDIM sampling~\cite{dhariwal2021diffusion,song2020denoising}. This technique is grounded in the hypothesis that the ordinary differential equation (ODE) process is reversible, especially in scenarios involving small step sizes.
The diffusion process is executed in reverse, meaning the transition is from $z_0$ to $z_T$, contrary to the typical $z_T$ to $z_0$ progression:
\begin{eqnarray}
\lefteqn{z_{t+1} = \sqrt{\frac{\alpha_{t+1}}{\alpha_t}} \, z_t} \nonumber\\ 
& & \quad{}+\left(\sqrt{\frac{1}{\alpha_{t+1}}-1}-\sqrt{\frac{1}{\alpha_t}-1}\right)\varepsilon_\theta\!\left(z_t, t, \mathcal{C}\right) .
\end{eqnarray}
Here, $\varepsilon_\theta$ is a noise predictor including an embedding of a text condition $\mathcal{C}$, while $z_0$ is the encoding of the provided real image. A guidance scale parameter $w$ is used to blend between a noise predictor with no text condition ($w = 0$) and $\varepsilon_\theta$ with $\mathcal{C}$.

To address the inefficiency of mapping each noise vector to a single image, we start with a default DDIM inversion at $w = 1$ as the pivot trajectory. Subsequently, we optimize around this trajectory using a standard guidance ratio of $w > 1$. In practical applications, individual optimizations are conducted for each step $t$ during the diffusion process, aiming to closely approximate the initial trajectory $z^*$:
\begin{equation}
\min \left\|z_{t-1}^*-z_{t-1}\right\|_2^2 \,,
\end{equation}
where $z_{t-1}$ represents the intermediate result of the optimization. We thus substitute the default blank text embedding with an optimized embedding. 
This is because the generated results are significantly influenced by the unconditional prediction \cite{mokady2023null}.

\section{Experiments}

We have compared our results with traditional methods such as `leave blank' and `stretch'. Additionally, we have selected the 3D Photography techniques of Shih et al.~\cite{shih20203d} for comparison, as well as the RePaint method of Lugmayr et al.~\cite{lugmayr2022repaint}, which involves using Stable Diffusion for inpainting images processed by the traditional `leave blank' method. 
It is important to emphasize that RePaint is not inherently designed for generating stereo image pairs. However, we believe that employing inpainting techniques to fill in the blank areas after creating stereo images is a very straightforward and common approach. Thus, we have chosen to compare with the latest model that achieves good results in various metrics within the same Stable Diffusion framework. This comparison is intended to demonstrate the innovation and advantage of our method.

\subsection{Quantitative evaluation} \label{sect:qe}

Since currently no metrics exists specifically for the comparison of stereo image pair generation, we quantitatively evaluate our results using the Middlebury~\cite{scharstein2014high} and KITTI~\cite{menze2015object} datasets. We evaluate the performance by generating the right-side image from the left-side image and its disparity map. We then compare the model-generated right-side image with the ground truth image. We calculated the Peak Signal-to-Noise Ratio (PSNR), Structural Similarity Index Measure (SSIM), and Learned Perceptual Image Patch Similarity (LPIPS) between the generated image and the ground truth. The results are in Table~\ref{tab:qe_mb} with some visuals in Figure~\ref{fig:comp_ssim}.
We provide the settings used for each method, more comparison figures, and a detailed explanation of the quantitative evaluation results in the supplemental document.

In Table~\ref{tab:time_cost}, we compare the time consumption of different methods for generating a single stereo image pair using an NVIDIA RTX3090 graphics card. Our method offers the capability to quickly generate high-quality stereo image pairs with direct integration into Stable Diffusion.

\begin{table*}[htbp]
    \centering
    \caption{Ablation study on Middlebury and KITTI.
    In the `Disparity Map' column, `GT' and `Pseudo' respectively indicate the use of groundtruth disparity maps or disparity maps generated by a Depth Estimation Model. In the `Technique Applied' column, `Attn Layer,' `SPSMD,' and `Deblur' represent the use of Self-Attention Layers Modification, Symmetric Pixel Shift Masking Denoise, and Deblur techniques, respectively. The symbol `\checkmark' denotes the adoption of these respective techniques. Bold numbers represent the best scores for that column. When employing Attn Layer and SPSMD together, LPIPS has a better score, but the effect of Deblur varies from image to image. When the LPIPS scores are comparable, the higher SSIM score indicates the better similarity. An example is shown in Figure~\ref{fig:ablation_sample_mb}.}
    \label{tab:ablation_mb}
    \vspace{-1ex}
    %\resizebox{\linewidth}{!}{
    {\small
    \resizebox{\linewidth}{!}{
    \begin{tabular}{cc|ccc|ccc|ccc}
        \hline
        \multicolumn{2}{c|}{\rule{0pt}{2.2ex}Disparity map}&\multicolumn{3}{c|}{Technique Applied} & \multicolumn{3}{c|}{Middlebury}& \multicolumn{3}{c}{KITTI} \\
        \hline
        \rule{0pt}{2.2ex}GT&Pseudo&Attn Layers&SPSMD&Deblur& PSNR $\uparrow$& SSIM $\uparrow$& LPIPS $\downarrow$&  PSNR $\uparrow$& SSIM $\uparrow$& LPIPS $\downarrow$\\
        \hline
        \rule{0pt}{3ex}&\checkmark&&&&$16.352^{+2.330}_{-2.139}$&$0.514^{+0.241}_{-0.169}$&$0.378^{-0.149}_{+0.181}$&$15.286^{+5.389}_{-4.622}$&$0.476^{+0.229}_{-0.346}$&$0.364^{-0.206}_{+0.265}$\\[3pt]
        &\checkmark&&\checkmark&&$\textbf{17.076}^{+4.450}_{-3.652}$&$0.549^{+0.190}_{-0.174}$&$0.191^{-0.073}_{+0.082}$&$15.305^{+6.013}_{-4.772}$&$0.474^{+0.228}_{-0.303}$&$0.230^{-0.115}_{+0.114}$ \\[3pt]
        &\checkmark&\checkmark&&&$15.421^{+2.849}_{-3.657}$&$0.478^{+0.245}_{-0.199}$&$0.255^{-0.123}_{+0.116}$&$\textbf{15.868}^{+7.775}_{-5.068}$&$\textbf{0.483}^{+0.240}_{-0.301}$&$0.212^{-0.113}_{+0.114}$\\[3pt]
        \checkmark&&\checkmark&\checkmark&&$15.456^{+2.669}_{-3.313}$&$0.468^{+0.252}_{-0.205}$&$0.231^{-0.088}_{+0.096}$ & $15.679^{+5.888}_{-5.486}$&$0.481^{+0.245}_{-0.310}$&$0.205^{-0.099}_{+0.135}$ \\[3pt]
        \checkmark&&\checkmark&\checkmark&\checkmark&$15.149^{+2.769}_{-3.174}$&$0.444^{+0.263}_{-0.231}$&$0.234^{-0.097}_{+0.117}$&$15.360^{+5.787}_{-5.332}$&$0.461^{+0.250}_{-0.306}$&$\textbf{0.200}^{-0.092}_{+0.127}$\\[3pt]
        &\checkmark&\checkmark&\checkmark&\checkmark&$16.753^{+4.815}_{-3.783}$&$0.540^{+0.197}_{-0.169}$&$0.174^{-0.071}_{+0.066}$&$15.269^{+7.923}_{-5.053}$&$0.458^{+0.261}_{-0.295}$&$0.204^{-0.111}_{+0.106}$\\[3pt]
        &\checkmark&\checkmark&\checkmark&&$16.980^{+4.737}_{-3.818}$&$\textbf{0.551}^{+0.208}_{-0.166}$&$\textbf{0.173}^{-0.069}_{+0.074} $&$15.589^{+8.061}_{-5.016}$&$0.479^{+0.241}_{-0.300}$&$ 0.209^{-0.116}_{+0.114}$ \\[1ex]
        \hline
        \\[-4ex]
    \end{tabular}}}
\end{table*}

\begin{table}
    \centering
    \caption{Time cost for different methods in seconds. We measure the total time consumed for each usage scenario, including the time taken to generate the images using Stable Diffusion. The scenarios are text to stereo image (T2SI), depth to stereo image (D2SI), and image to stereo image (I2SI). The time in parenthesis is the cost excluding the time spent on generating images with Stable Diffusion. For D2SI, our method, being directly integrated into Stable Diffusion, requires only a single pass of sampling to generate stereo image pairs. In I2SI, our method requires the use of null-text inversion~\cite{mokady2023null} for $x_t$, resulting in an extra 23 seconds. %of time expenditure.
    }
    \label{tab:time_cost}
    \vspace{-1ex}  
    %{\small
    \begin{tabular}{@{\,}c|ccc@{\,}}
        \hline
         \rule{0pt}{2.2ex}Methods& T2SI & D2SI & I2SI\\
         \hline
         \rule{0pt}{2.2ex}3D Photography \cite{shih20203d}& 245  (231) & 247  (231)  & 231  \\
         Repaint\cite{lugmayr2022repaint}& 338  (324) & 340  (324)  & 324 \\
         Ours& 32 (18) & 18  & 40  (17)\\
         \hline
    \end{tabular}
\end{table}

\begin{figure}
    \centering
    \includegraphics[width=\linewidth]{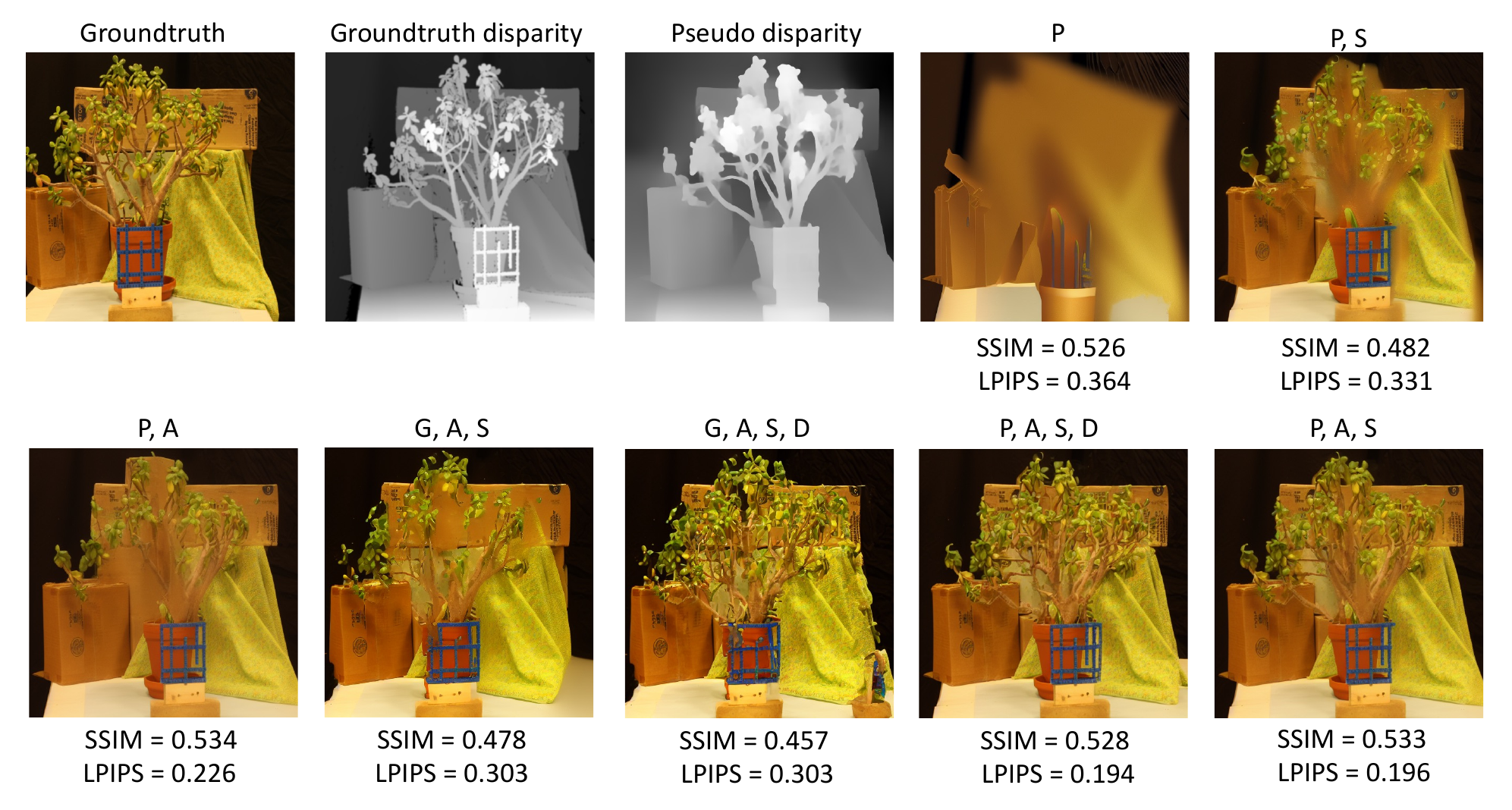} \\[-1ex]
    \includegraphics[width=\linewidth]{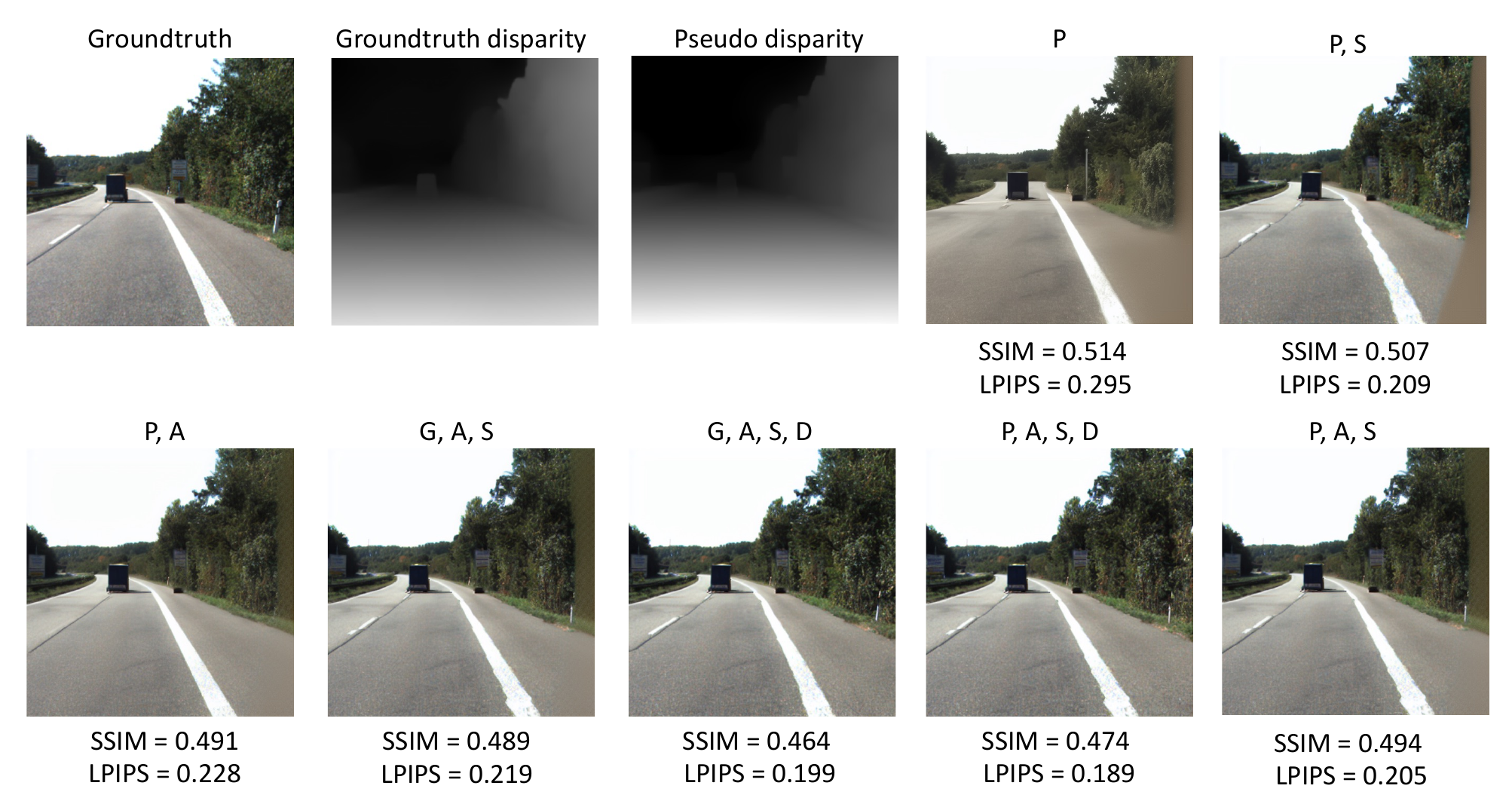}\\[-2ex]
    \caption{
    Ablation example of Middlebury (up) and KITTI (down). 
    In the images, `P' and `G' respectively denote whether the image has been guided by a Pseudo disparity map or a Groundtruth disparity map. `A', `S', and `D' indicate the use of Attention layers modification, Symmetric Pixel Shift Masking Denoise, and Deblur technique, respectively. The lower scores associated with the use of Groundtruth disparity maps in Middlebury may be attributed to their generally higher precision and complexity. This heightened detail can render pixel shift operations during image generation more intricate and sensitive. Our Stereo Pixel Shift operation is executed within a smaller latent space (64×64), where minor pixels, such as those around tree trunks and leaves, might be overlooked. In contrast, disparity maps generated by depth estimation models, with their lower precision, are more conducive to Pixel Shift in the latent space without sacrificing image detail.
    }
    \label{fig:ablation_sample_mb}
\end{figure}

\subsection{User evaluations}
\begin{figure}
    \centering
    \vspace{-0.8ex}
    \includegraphics[width=0.7\linewidth]{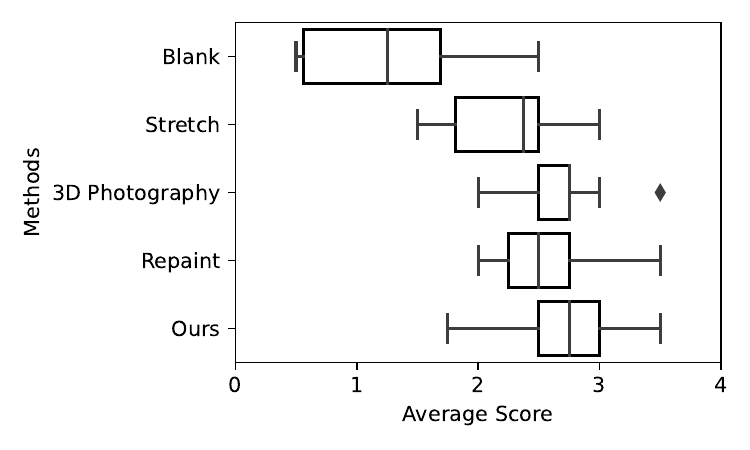} \\[-2ex]
    \caption{User evaluation results. Distribution of the scores provided by users on the test scenes.\vspace{-2ex}}
    \label{fig:userscores}
\end{figure}

In our user tests, we adopted a more practical and user-centric approach. The users' input were text prompts used to generate stereo image pairs using Stable Diffusion. For benchmarking, we compared this with other methods by generating the left-side images using Stable Diffusion, obtaining the corresponding disparity maps via a depth estimation model, and then using the respective methods to generate stereo image pairs.
We utilized Google Cardboard and presented the stereo images on mobile phones, inviting participants to assess the image quality and correctness of the 3D perception. Ratings ranged from 0 to 5, with 5 being the highest and 0 being the lowest. Some test pictures are shown in Figure~\ref{fig:comparison} of the supplemental document.

The results of the user tests showed that our method has the highest average but it did not significantly outperform the others. This was anticipated, as when viewing stereo images, people tend to focus more on the overall image rather than the details. In terms of ease of use, our proposed method has a clear advantage. It is simpler, does not require an additional inpainting model, and can be seamlessly integrated with Stable Diffusion.

\subsection{Ablation study} \label{sect:ablation}

We conducted ablation studies on the proposed method to evaluate the impact of images guided by either Groundtruth disparity maps or Pseudo disparity maps (generated by a depth estimation model), as well as the effects of using Symmetric Pixel Shift Masking Denoise, Attention Layer Modification, and Deblur techniques on PSNR, SSIM, and LPIPS scores. The results are shown in Table \ref{tab:ablation_mb}. Figure~\ref{fig:ablation_sample_mb} presents a visual representation of an example from the Middlebury dataset and KITTI to intuitively demonstrate the impact of each factor on the image generation outcomes, explaining the reason that scores using Groundtruth disparity maps in the Middlebury dataset are unexpectedly lower than those using Pseudo disparity maps.

\section{Limitations and Discussion}

Our method relies on the disparity map. If the results generated by other depth estimation models are inaccurate, our method will also be unable to produce high-quality stereo images. Furthermore, when using high-precision disparity maps obtained from device measurements, the results may not be entirely satisfactory, as shown in Table~\ref{tab:ablation_mb} and Figure~\ref{fig:ablation_sample_mb}.

\begin{figure}
    \centering
    \includegraphics[width=1\linewidth]{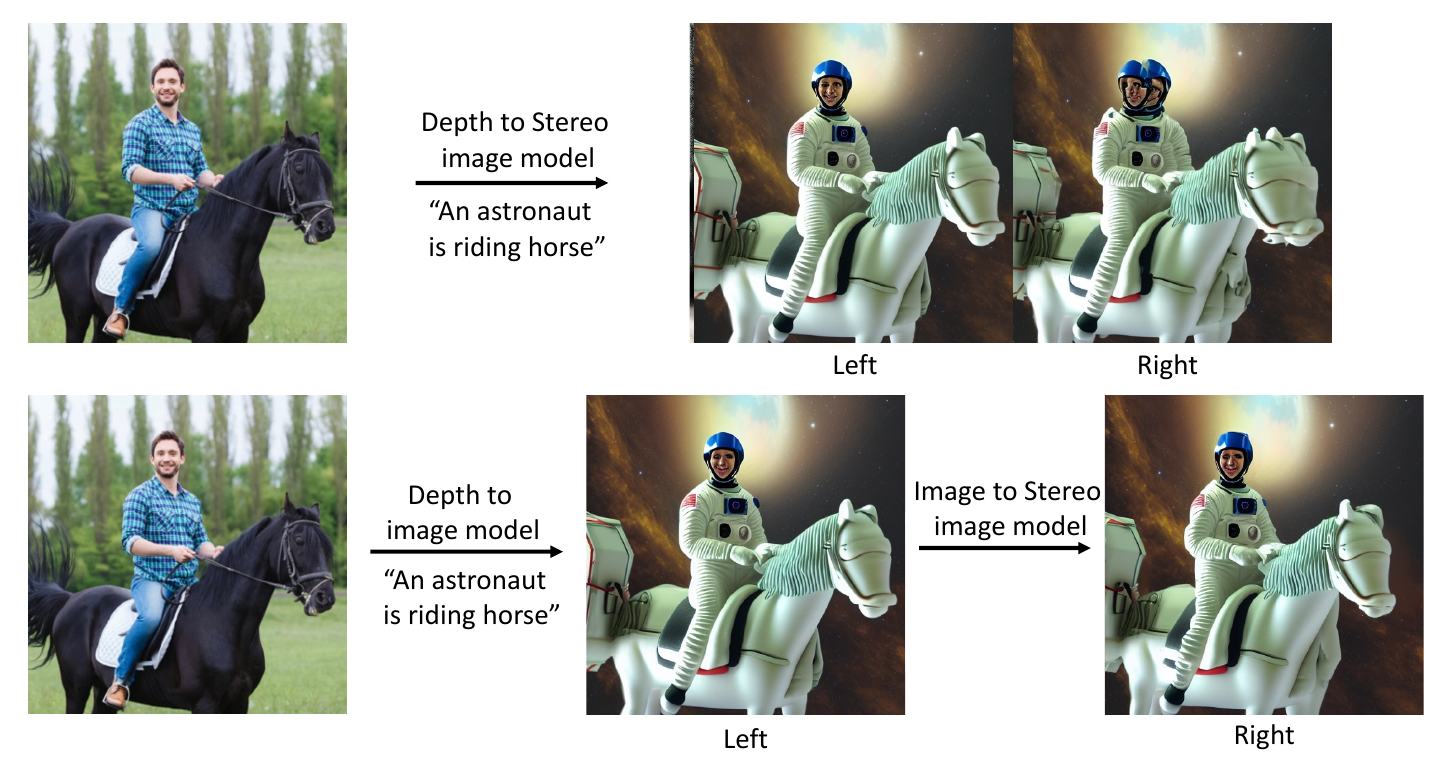} \\[-2ex]
    \caption{
    Limitation: Generating compositionally similar stereo images directly from a disparity map may sometimes fail. However, this issue can be mitigated by first generating a compositionally similar left image using the disparity map, and then employing the Image to Stereo Image method to generate the right image. This two-step process helps avoid such failures.
    %\vspace{-2ex}
    }
    \label{fig:limit}
\end{figure}

When using our depth to stereo image model, one may observe overlapping areas in the generated images. This issue might stem from the \textit{LatentDepth2ImageDiffusion} model we used, which tends to fill blank areas with pixels from adjacent main subjects rather than background elements. In such cases, a better-quality image can be generated by first generating a single image using the \textit{Depth2Image} model, and then applying our Image to Stereo Image Pairs method, as illustrated in Figure~\ref{fig:limit}.

\begin{figure}
    \centering
    \includegraphics[width=0.9  \linewidth]{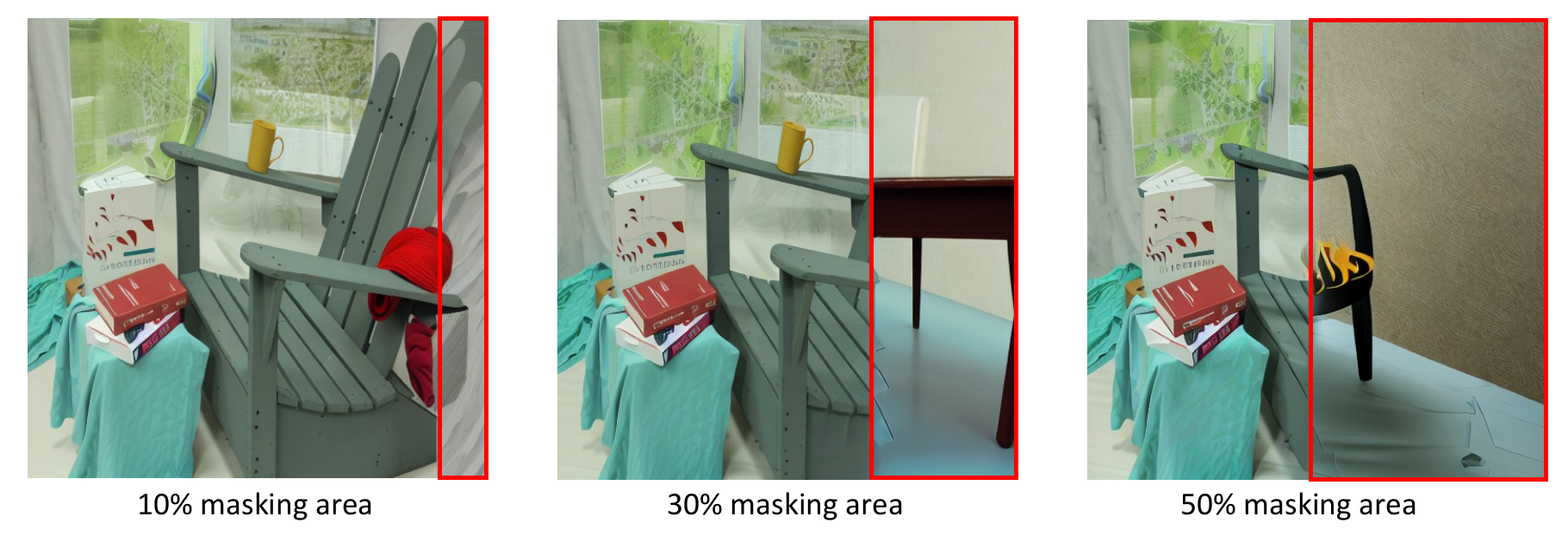} \\[-2ex]
    \caption{
    Tests for inpainting tasks using our proposed method, the red-colored areas represent the masked regions.
    %\vspace{-2ex}
    }
    \label{fig:discuss_inpaint}
\end{figure}

We found that our method can be used for inpainting tasks with the original text prompt to a Stable Diffusion image model. We conducted a simple test where, after obtaining $ \bm{x}_t $ using null-text inversion, we applied various masking ratios to the right side and tested whether Stable Diffusion could fill in the blank areas within the mask during denoising. The results, as shown in Figure~\ref{fig:discuss_inpaint}, indicate that our method is somewhat effective for inpainting when a smaller area of the image is masked. However, when a larger portion of the image is masked, the inpainting results exhibit a strong patchwork appearance. Applying our method to inpainting tasks might require further modifications to both the model and the technique.

\section{Conclusion}

We proposed a novel method for generating stereo image pairs by modifying the latent vector of Latent Stable Diffusion. We implement Stereo Pixel Shift on the left latent vector and its corresponding disparity map, and during the denoising process, we ensure consistency between the left and right images through Symmetric Pixel Shift Masking Denoise and Self-Attention Layer Modification. Our approach differs fundamentally from traditional inpainting pipelines and can be seamlessly integrated into existing Stable Diffusion models, offering end-to-end capabilities for text prompt to stereo image, depth to stereo image, and image to stereo image generation, all without the need for fine-tuning any parameters and using only the original Stable Diffusion model. Our method achieved better scores on both the KITTI and Middlebury datasets.

%% file: sec/X_suppl.tex
\clearpage
\setcounter{page}{1}
\maketitlesupplementary

\section{Quantitative evaluation experiments setting}
In this section, we will provide a detailed description of the settings for each method.
3D Photography does not provide a direct method for generating stereo image pairs, its output is a mesh, which requires rendering to obtain images. Therefore, we manually set the left and right camera matrices as follows:
\[ M_{\text{left}} = \begin{bmatrix} 1 & 0 & 0 & 0 \\ 0 & 1 & 0 & 0 \\ 0 & 0 & 1 & 0 \\ 0 & 0 & 0 & 1 \end{bmatrix} \,,\quad
 M_{\text{right}} = \begin{bmatrix} 1 & 0 & 0 & -0.04 \\ 0 & 1 & 0 & 0 \\ 0 & 0 & 1 & 0 \\ 0 & 0 & 0 & 1 \end{bmatrix}\,. \]
After rendering with these settings, we obtain the left and right images of the stereo image.

Given that Stable Diffusion can only generate images of $512 \times 512$ resolution, and the Middlebury dataset images are about 5 million pixels, we scaled both the dataset images and the corresponding depth maps to $512 \times 512$. For the Middlebury dataset, whose groundtruth disparity maps are noisy, we applied a Gaussian blur with a radius of 3 to smooth the disparity maps. 
Regarding the KITTI dataset, where the image size is $375 \times 1242$ with an aspect ratio of approximately 3.3, directly scaling images to $512 \times 512$ could lead to excessive stretching, negatively impacting many models' performance. Therefore, we proportionally scaled the images to $512 \times 1696$ and then applied a center crop to $512 \times 512$. Because a null-text inversion technique is required, we used the Stable Diffusion version 1.5 for this test, setting the denoising steps to 50.

For the 3D photography method~\cite{shih20203d}, we used the disparity map generated by the integrated MiDaS model~\cite{Ranftl2022} within its framework instead of the groundtruth disparity map. This was due to the extensive time required—up to two hours—for mesh reconstruction of a single image using the groundtruth disparity map with 3D photography. We hypothesize that this inefficiency arises when 3D photography attempts to reconstruct stereo image pairs from the disparity map, necessitating operations like breaking up discontinuous vertices in the mesh. Such processes become computationally intensive when the groundtruth disparity map is excessively noisy, leading to a proliferation of isolated vertices that consume substantial CPU resources. 
For the purpose of benchmarking and considering the rarity of obtaining groundtruth disparity maps in practical scenarios, we evaluated the results using both groundtruth disparity maps (denoted as GT disparity) and pseudo disparity maps generated by depth estimation models (denoted as Pseudo disparity). The depth estimation model we employed was DPT~\cite{Ranftl2021}. Since the use of Deblur results in lower scores, neither method employed deblur; details can be found in Sec.~\ref{sec:ablation}.
When creating stereo images using RePaint~\cite{lugmayr2022repaint}, we generate a mask for the blank areas left after moving the left-side image and then perform inpainting on the masked areas. The \textit{inet256} model was utilized for this purpose, and was trained on ImageNet. Since RePaint's maximum supported output image size is $256 \times 256$, we downsized the images to $256 \times 256$ before conducting inpainting. However, considering that all other methods are evaluated at a $512 \times 512$ resolution, for fairness, we only upscaled the inpainted area within the mask from $256 \times 256$ to $512 \times 512$, while maintaining the original resolution for the area outside the mask.

\section{Analysis of Quantitative evaluation results}

\begin{figure*}
    \centering
    \includegraphics[width=\linewidth]{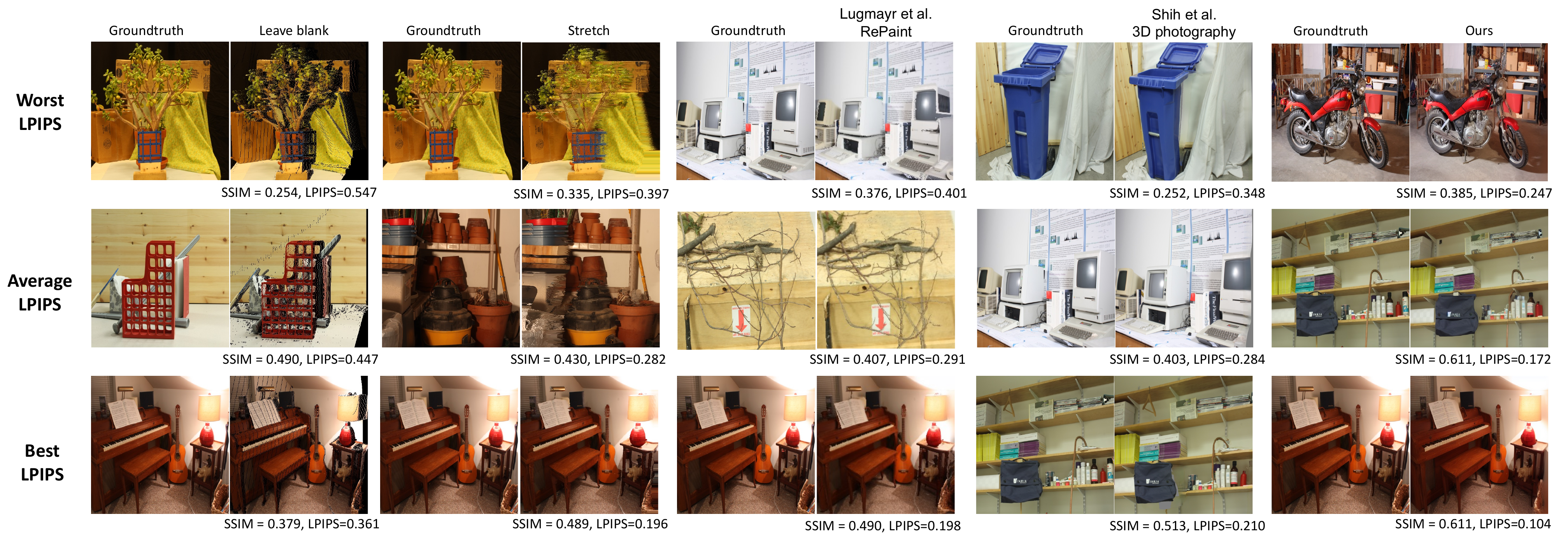} \\[-2ex]
    \caption{
    Comparing different methods by Perceptual Image Patch Similarity (LPIPS) scores. We evaluate the right-side images generated from left-side images and disparity maps using various methods: 'Worst LPIPS', 'Average LPIPS', and 'Best LPIPS'. These represent, respectively, the images with the highest (worst) LPIPS score, the image closest to the average LPIPS score, and the image with the lowest (best) LPIPS score for each method. We also annotate each image with its Structural Similarity Index Measure (SSIM) for reference. 
    }
    \label{fig:comp_lpips}
\end{figure*}
The use of the null-text inversion~\cite{mokady2023null} technique inherently causes distortion in images. On the Middlebury dataset, reference scores (for images generated by Stable Diffusion to be the same as the input) are: PSNR = 27.967, SSIM = 0.847, LPIPS = 0.046. The reference scores for the KITTI dataset are: PSNR = 25.615, SSIM = 0.762, LPIPS = 0.072. These scores represent the best possible outcomes achievable with the method we proposed.
The quantitative analysis results, as seen in Table~\ref{tab:qe_mb}, indicate that our proposed method achieves state-of-the-art scores on both the datasets. Furthermore, as illustrated in Fig.~\ref{fig:comp_lpips}, we selected images representing the best LPIPS, those closest to the average LPIPS, and the worst LPIPS from each method. This selection was made to visually demonstrate the differences in images generated by each method. Fig.~\ref{fig:comp_ssim} showcases images with the lowest SSIM, closest to the average SSIM, and the highest SSIM scores when using our method. We also include a comparison to the outcomes when other methods are applied to the same images. We have also magnified some details to facilitate an intuitive comparison of the primary methods.

We also noted that the scores for the KITTI dataset are lower compared to those of the Middlebury dataset. 
However, if we convert the best scores into percentages relative to the Stable Diffusion reference scores, the results are as follows. For the Middlebury dataset, with SSIM = 0.551, it is 65.1\% of the best score of 0.847, and for LPIPS = 0.173, the reference score of 0.046 constitutes 26.6\% of the best score of 0.173 (the higher the percentage, the better). Similarly, for the KITTI dataset, SSIM is 63.1\% of the reference score of 0.762, and the reference score for LPIPS of 0.072 is 35.1\% of the best score. The model actually performs better on the KITTI dataset in terms of LPIPS.
Another possible reason for this is the larger baseline distance $ B $ of the cameras used to capture the KITTI dataset images, which in turn requires a larger scale factor $ s $  (KITTI $ s = 20 $ , Middlebury $ s = 9 $). This larger scale factor means that, when generating stereo image pairs, the corresponding pixels in the KITTI dataset images have to move a greater distance, resulting in more extensive blank areas.

\section{Analysis of Ablation}

Deblur has a certain negative impact on LPIPS and SSIM scores on Middlebury dataset, with a more pronounced effect on SSIM. This is because blurred images contain fewer high-frequency details, implying less noise and finer details. Since SSIM focuses more on large-scale structural features at lower frequencies, these features might appear more pronounced and consistent in blurred images, leading to higher SSIM scores. Unlike traditional metrics like SSIM or PSNR, LPIPS emphasizes perceptual differences rather than just pixel-level discrepancies, hence the lesser impact of Deblur on LPIPS scores. A lower LPIPS score with highter SSIM scores indicates closer approximation to the original image.

On the KITTI dataset, the scores for Groundtruth and Pseudo disparity maps are more aligned with general expectations. Compared to the high-precision and complex Groundtruth disparity maps in the Middlebury dataset, the Groundtruth disparity maps in the KITTI dataset are relatively straightforward, mostly depicting driving scenes. Therefore, stereo images guided by Groundtruth disparity maps scored higher than those guided by Pseudo disparity maps. We believe that the positive effect of Deblur in the KITTI data set is due to the large scale factor $s$, which makes the larger blank area left after Pixel shift unable to be filled during denoise.
It's also important to note that the LPIPS score is a better indicator of the overall similarity of images. Therefore, a higher SSIM score accompanied by a higher LPIPS score does not necessarily imply a greater similarity to the original image, as demonstrated in Fig.~\ref{fig:ablation_sample_mb}. However, when the LPIPS scores are comparable, the SSIM score becomes a more effective measure for assessing the similarity of images.

\section{Attention module modification details}

%Attn mechanism
Within the Stable Diffusion model, the denoising U-Net is structured as a series of basic blocks. Each basic block incorporates a residual block, a self-attention module, and a cross-attention module which can be represented as~\cite{yang2022diffusion,song2020denoising,preechakul2021diffusion}.
\begin{equation}
\mathrm{Attention}(Q, K, V)=\mathrm{Softmax}\left(\frac{Q K^\top}{\sqrt{d}}\right) V \,,
\end{equation}
where $Q$ represents the query, while $K$ and $V$ represent the key and value, respectively, and $d$ is the output dimension of the key and query features. The values are obtained through linear projection. When there is an input context, it functions as cross-attention. In the absence of context, it operates as self-attention. Cross-attention is commonly employed in tasks involving text-guided image editing~\cite{hertz2022prompt,cao2023masactrl}.

In the case of self-attention, non-rigid editing cannot be performed as the semantic layout and structures are maintained. Similar to sharing semantic information between different samples in the same batch using 3D convolution to align content across batches in video generation tasks~\cite{wang2023videocomposer,blattmann2023align,guo2023animatediff}, applying self-attention between samples within the same batch has a comparable effect~\cite{wu2022tune,chen2023control}. 
Querying the left-side image using the key and value of the right-side image in a unidirectional manner, enhancing the alignment from right image to left, is termed unidirectional self-attention. In contrast, employing queries from both the left and right sides to mutually query each other is referred to as bidirectional self-attention. 
However, bidirectional self-attention has a significant drawback: it aligns the left and right images with each other, thereby altering the input left-side image. Although this can enhance alignment, it is not a suitable option when users wish to keep the input image unchanged. Thus, despite its potential to improve alignment, the bidirectional approach may not be preferable if it is crucial to maintain the integrity of the input image. The algorithm is explained in the appendix.

We apply this attention control to all layers of the U-Net to achieve the best alignment results. Although another study observed that applying attention control to all layers results in exactly the same images~\cite{cao2023masactrl}, in our method, stereo shifts have already been applied, which leads to content consistency while the main subject is shifted to different positions, precisely the outcome we desire.

\section{Attempts of fine-tuning the Stable Diffusion model to generate stereo image pairs}

\begin{figure}[htbp]
    \centering
    \includegraphics[width=0.6\linewidth]{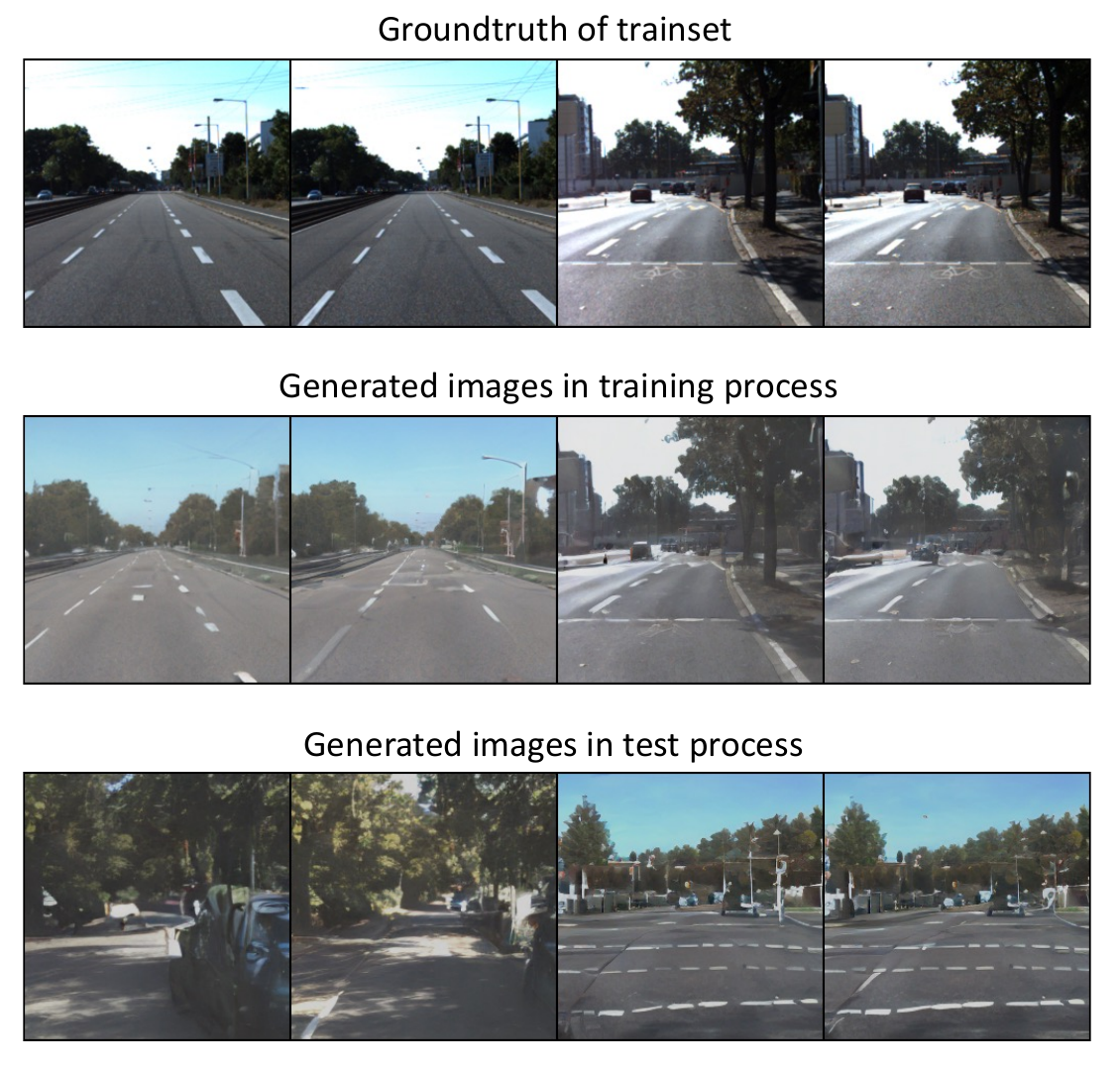} \\[-2ex]
    \caption{
    Example of images generated by stereo fine-tuned Stable Diffusion: The images reveals that while the generated left and right images exhibit certain similarities, the extent of this resemblance falls significantly short of the requirements for stereo imaging. Even during training, maintaining pixel-level consistency between the left and right images proves challenging, and the quality of images generated during tests exhibits notable deficiencies.
    }
    \label{fig:finetune}
\end{figure}

In this section, we briefly present our initial attempts at fine-tuning Stable Diffusion for generating stereo image pairs. This approach was unsuccessful in producing high-quality stereo image pairs. 

ControlNet~\cite{chen2023control}, known for its capability to manipulate the posture of images generated by Stable Diffusion, produces images that are structurally similar to the input image but with different content. We hypothesized that this might be beneficial for generating stereo images. Consequently, we adopted an architecture similar to ControlNet. A neural network block \( F(\cdot; \Theta) \) with a set of parameters \( \Theta \) transforms a feature map $\boldsymbol{x}$ into another feature map \( \boldsymbol{y} \).
\begin{equation}
    \boldsymbol{y}=F (\boldsymbol{x}; \Theta) \,.
\end{equation}
We have frozen all the parameters $\Theta$ of the original Stable Diffusion model and created a trainable copy $\Theta_c$. The neural network blocks are interconnected through a distinctive convolution layer, which is initialized with zero weights and biases. The operation can be represented by the following equation
\begin{equation}
\boldsymbol{y}_{\mathrm{c}}=\mathcal{F}(\boldsymbol{x} ; \Theta)+\mathcal{Z}\left(\mathcal{F}\left(\boldsymbol{x}+\mathcal{Z}\left(\boldsymbol{c} ; \Theta_{\mathrm{z} 1}\right) ; \Theta_{\mathrm{c}}\right) ; \Theta_{\mathrm{z} 2}\right) ,
\end{equation}
where \( y_c \) represents the output of this neural network block. The operation \( Z(\cdot; \cdot) \) denotes a zero convolution operation, and \( \{\Theta_{z1}, \Theta_{z2}\} \) represents two instances of parameters, each corresponding to a distinct instance of the zero convolution operation.

Using ControlNet only maintains the general content of the images, which is insufficient for generating stereo image pairs. We aim for Stable Diffusion to generate stereo image pairs concurrently. To achieve this, we align even-numbered images in the batch with their adjacent odd-numbered counterparts, such as 0 with 1, and 1 with 2, to create a stereo effect between each adjacent pair. Inspired by VideoLDM\cite{blattmann2023align}, we introduce a 3D convolution layer and a temporal attention layer into the Stable Diffusion architecture. These layers are added after Stable Diffusion's existing spatial layers in the U-Net. The function of 3D convolution layers is to break the information isolation between different samples in the same batch. Before feeding the intermediate features to the 3D convolution layer, we reshape the features from \texttt{[b c h w]} to \texttt{[b/2 2 c h w]}, where \texttt{b}, \texttt{c}, \texttt{h}, \texttt{w} represent batch size, color channel, height, and width, respectively. The \texttt{2} in the reshaped second item represents the left and right images, allowing the newly added 3D convolution block to learn the distribution of the left and right stereo image pairs. The structure of the temporal attention layer is same as that in Stable Diffusion, assisting the 3D convolution layer in distinguishing different timesteps during the denoise process. 

However, the use of ControlNet combined with 3D convolution layers is still insufficient to generate stereo image pairs. Despite a certain degree of consistency between the left and right images, the main objects within these images do not maintain a strict correspondence. For example, a car appearing in the center of the left image may appear in a considerably random position in the right image. Although the KITTI dataset is captured with the same devices and, in theory, 3D convolution blocks should be able to learn the devices' parameters and estimate the displacement of objects in the right image relative to the left, this proves to be quite challenging in practice. Hence, we introduced a disparity map as an additional condition. Our purpose was to use the disparity map of the left image as guidance to assist the 3D convolution blocks in estimating the pixel displacement in the right image. Using the disparity map as an additional condition for Stable Diffusion significantly improved the quality of the generated images, but the detail quality still did not meet our standards. Even when limiting the generation type to driving scenes, the probability of producing flawed images remained high. Therefore, we abandoned this approach.
Fig. \ref{fig:finetune} shows the example of images generated using fine-tuned Stable Diffusion.

\section{Ablation of Bidirectional attention and Stereo Pixel Shift}
\begin{figure}[htbp]
    \centering
    \includegraphics[width=0.5\linewidth]{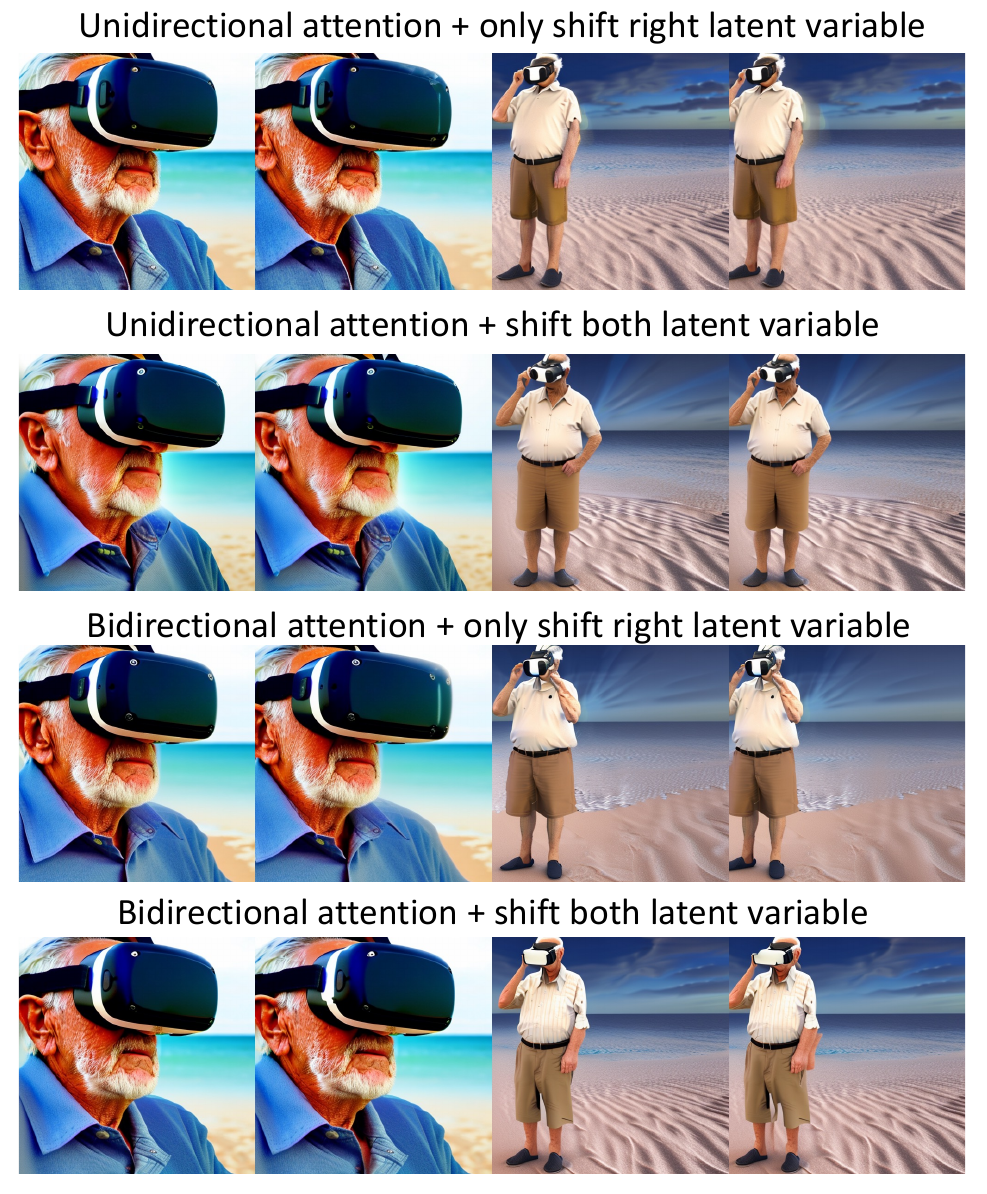} \\[-1ex]
    \caption{
    Ablation of Bidirectional attention and Stereo Pixel Shift: The implementation of Bidirectional attention and the simultaneous application of Stereo Pixel Shift to the left and right latent variables theoretically enhances the consistency between the two images. However, this approach may induce certain changes in the original images, which are currently uncontrollable.
    }
    \label{fig:ablation_app}
\end{figure}
Incorporating Bidirectional Attention and applying Stereo Pixel Shift to both the left and right latent variable can alter the original image, making it unsuitable for quantitative analysis. Therefore, we only partially showcase the results of the text prompt to stereo image generation, as depicted in Fig.~\ref{fig:ablation_app}. The simultaneous application of Bidirectional Attention and Stereo Pixel Shift to both left and right latent variable may induce changes in the original image. These modifications are currently uncontrollable. However, this may suggest a new potential of our approach: a method of controlling the generated images, akin to ControlNet, but without the need for fine-tuning.

\section{User test images}

\begin{figure*} %[htbp]
    \centering
    \includegraphics[width=\linewidth]{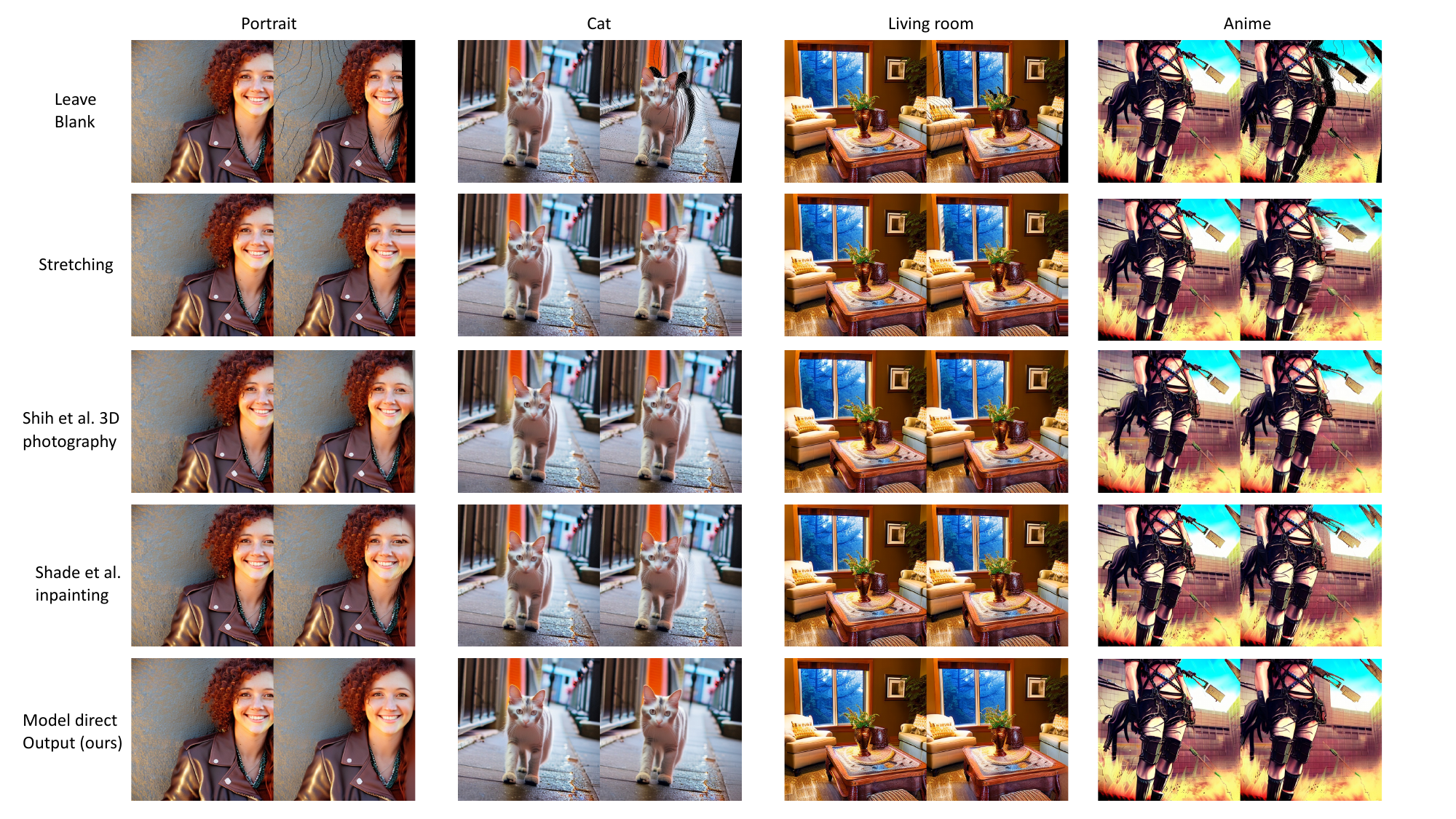} 
    \includegraphics[width=\linewidth]{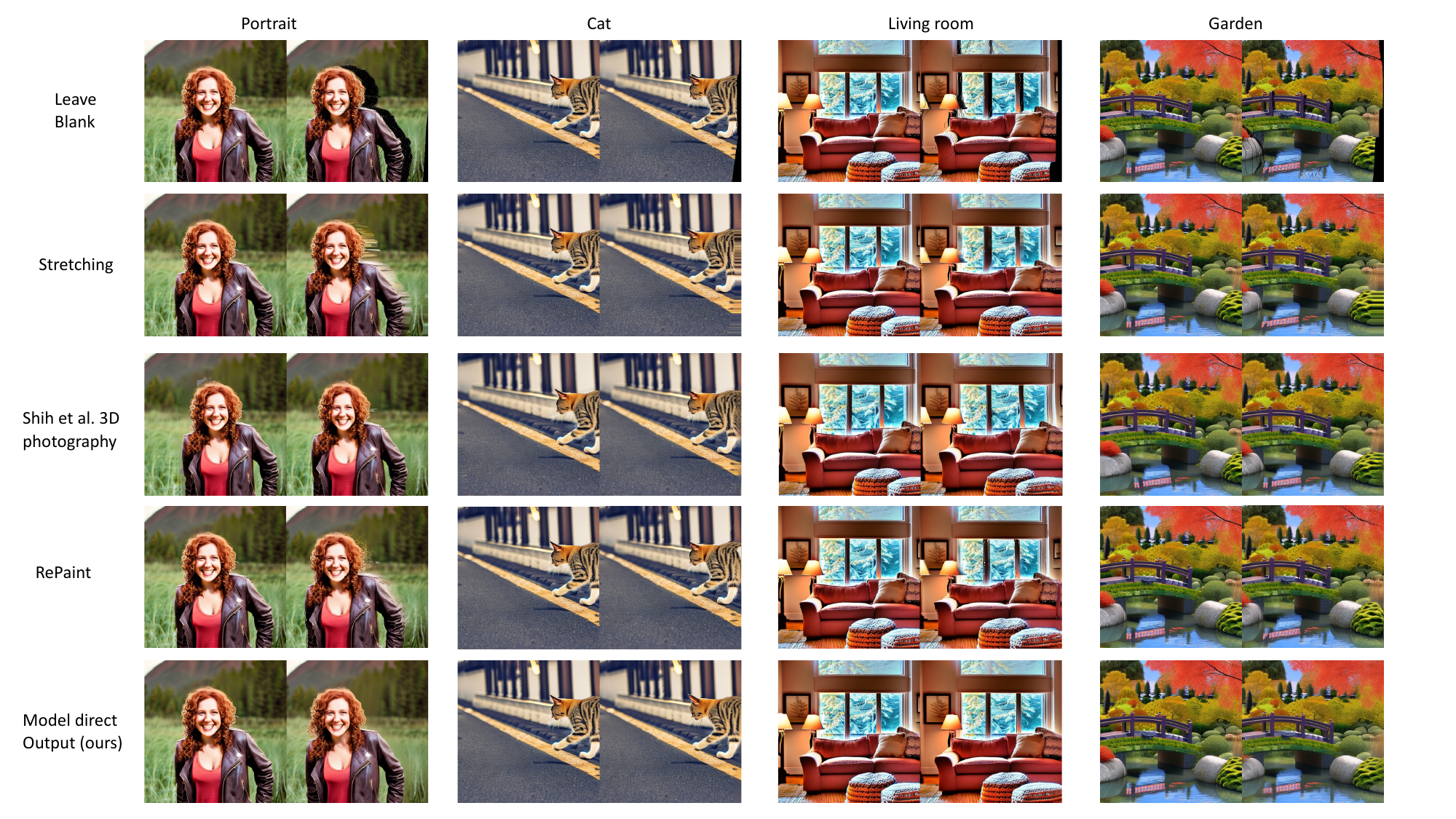} \\[-2ex]
    \caption{Comparison of stereo image generation techniques. RePaint et al.~\cite{lugmayr2022repaint} indicates using their inpainting model to fill the blank area. HINT: The images can be viewed using the autostereogram technique to achieve a 3D effect. (Keep your eyes steady and maintain an unfocused gaze, try adjusting the eyes' focus and the distance between the autostereogram and your eyes slightly.)}
    \label{fig:comparison}
\end{figure*}

In Fig.~\ref{fig:comparison}, we show the example images used for our user evaluation. 